\documentclass[11pt]{article}

\pdfoutput=1 
\usepackage[a4paper,hscale=0.725,vscale=0.78]{geometry}

\usepackage{amsmath}
\usepackage{amsfonts}
\usepackage{url}

\usepackage{mathptmx}

\usepackage[linkcolor=black,bookmarksopen]{hyperref}

\DeclareMathOperator*{\allargsmax}{allargsmax}
\DeclareMathOperator*{\pickargmax}{pickargmax}

\DeclareMathOperator*{\minop}{min}

\def\gnull{g\raisebox{-.4ex}{$_0$}}

\def\lpower{\text{\it lpower}}
\def\bjp{\text{\it bjp}}
\def\bpe{\text{\it bpe}}
\def\bnp{\text{\it bnp}}
\def\bjpl{\text{\it button\_just\_pressed}}
\def\bpel{\text{\it button\_pressed\_earlier}}
\def\bnpl{\text{\it button\_not\_pressed}}
\def\dntu{\text{\it dntu}}

\def\bif{\text{\bf if}\,}
\def\belse{\,\text{\mbox{\bf else}}\,}
\def\bfi{\,\text{\bf f\hspace{0.2ex}i}}

\def\mystack#1#2{\hspace*{-.5ex}\raisebox{-1ex}{$\genfrac{}{}{0pt}{1}{\mbox{#1}}{\mbox{#2\hfill}}$}}

\usepackage[font=it,labelfont=it,margin=10pt]{caption}

\newenvironment{smallenumerate}{
\vspace{-1.5ex}
\begin{enumerate} \setlength{\itemsep}{-.4ex} 
}{
\end{enumerate}
\vspace{-1.5ex}
}

\begin{document}
%%%!2
\title{\bf Corrigibility with Utility Preservation}
\author{{\bf Koen Holtman}\thanks{Permanent e-mail address: {\tt Koen.Holtman@ieee.org}}\\
{\normalsize Eindhoven, The Netherlands}}
\date{April 2020 (Version 2)}
\maketitle
\vspace*{-5ex}
\begin{abstract}
Corrigibility is a safety property for artificially intelligent
agents.  A corrigible agent will not resist attempts by authorized
parties to alter the goals and constraints that were encoded in the
agent when it was first started.  This paper shows how to construct a
safety layer that adds corrigibility to arbitrarily advanced utility
maximizing agents, including possible future agents with Artificial
General Intelligence (AGI).  The layer counter-acts the emergent
incentive of advanced agents to resist such alteration.

A detailed model for agents which can reason about preserving their
utility function is developed, and used to prove that the
corrigibility layer works as intended in a large set
of non-hostile universes.  The corrigible agents have an emergent
incentive to protect key elements of their corrigibility layer.
However, hostile universes may contain forces strong enough to break
safety features.  Some open problems related to graceful
degradation when an agent is successfully attacked are identified.

The results in this paper were obtained by concurrently developing an
AGI agent simulator, an agent model, and proofs.  The simulator is
available under an open source license. The paper contains simulation
results which illustrate the safety related properties of corrigible
AGI agents in detail.
\end{abstract}

\section{Introduction}%%%!1

In recent years, there has been some significant progress in the field
of Artificial Intelligence, for example \cite{silver2017mastering}.
It remains uncertain whether agents with Artificial General
Intelligence (AGI), that match or exceed the capabilities of humans in
general problem solving, can ever be built, but the possibility cannot
be excluded \cite{everitt2018agi}.  It is therefore interesting and
timely to investigate the design of safety measures that could be
applied to AGI agents.  This paper develops a safety layer for
ensuring corrigibility, which can be applied to any AGI agent that is
a utility maximizer \cite{von1944theory}, via a transformation on its
baseline utility function.

Corrigibility \cite{corr} is the safety property where an agent will
not resist any attempts by authorized parties to change its utility
function after the agent has started running.  The most basic
implication is that a corrigible agent will allow itself to be
switched off and dismantled, even if the agent's baseline utility
function on its own would create a strong incentive to resist this.

Corrigibility is especially desirable for AGI agents because it is
unlikely that the complex baseline utility functions built into such
agents will be perfect from the start.  For example, a utility
function that encodes moral or legal constraints on the actions of the
agent will likely have some loopholes in the encoding.  A loophole may
cause the agent to maximize utility by taking unforeseen and highly
undesirable actions. Corrigibility ensures that the agent will not
resist the fixing of such loopholes after they are discovered.  Note
however that the discovery of a loophole might not always be a
survivable event.  So even when an AGI agent is corrigible, it is
still important to invest in creating the safest possible baseline
utility function.  To maximize AGI safety, we need a layered approach.

It is often easy to achieve corrigibility in limited artificially
intelligent agents, for example in self-driving cars, by including an
emergency off switch in the physical design, and ensuring that any
actuators under control of the agent cannot damage the off switch, or
stop humans from using it. The problem becomes hard for AGI agents
that have an unbounded capacity to create or control new actuators
that they can use to change themselves or their environment. In terms
of safety engineering, any sufficiently advanced AGI agent with an
Internet connection can be said to have this
capacity. \cite{omohundro2008basic} argues that in general, any
sufficiently advanced intelligent agent, designed to optimize the
value of some baseline utility function over time, can be expected to
have an emergent incentive to protect its utility function from being
changed.  If the agent foresees that humans might change the utility
function, it will start using its actuators to try to stop the humans.
So, unless special measures are taken, sufficiently advanced AGI
agents are not corrigible.

\subsection{This paper}%%%!

The main contribution of this paper is that it shows, and proves
correct, the construction of a corrigibility safety layer that can be
applied to utility maximizing AGI agents.  It extends
and improves on previous work
\cite{corr} \cite{corra} in by resolving the issue of
utility function preservation identified in \cite{corr}.  The design
also avoids creating certain unwanted manipulation incentives
discussed in \cite{corr}.

A second contribution is the development of a formal approach for
proving equivalence properties between agents that have the emergent
incentive to protect their utility functions.  The construction of an
agent with a utility function preservation incentive that is boosted
beyond the emergent level is also shown.  Some still-open problems in
modeling agent equivalence and achieving graceful degradation in
hostile universes are identified.

A third contribution of this paper is methodological in nature.  The
results in this paper were achieved using an approach where an AGI
agent simulator, an agent model, and proofs were all developed
concurrently.  The development of each was guided by intermediate
results and insights obtained while developing the other.  We
simulate a toy universe containing an agent that is
super-intelligent
\cite{bostrom2014superintelligence}, in the sense that the agent is
maximally adapted to solving the problem of utility maximization in
its universe.  The methodology and tools developed may also be
useful to the study of other open problems in AGI safety.  The
simulator developed for this paper is available on GitHub
\cite{simgithub} under an Apache open source license.

Finally, this paper shows simulation runs that illustrate the behavior
of a corrigible agents in detail, highlighting implications relevant
for safety engineers and policy makers.  While some policy trade-offs
are identified, the making of specific policy recommendations is out
of scope for this paper.

\subsection{Introduction to Version 2}%%%!

The main change in this version 2 of the paper is that it makes
different choices in developing the mathematical notation.  Version 1
developed its notation to stay close to that in \cite{corr}. This
version 2 adapts the Markov Decision Process (MDP) related notation
conventions.  The intention is to make the formalism and proofs
developed more accessible to a larger audience, even though at a
deeper level, they remain identical to those in version 1.

This version 2 also fixes several grammatical and typographical errors
in the text and proofs of version 1.  To improve the order or
presentation, some material was moved to appendices
\ref{sabotage} and \ref{secondproof}. The text in Appendix
\ref{annex}, on the interpretation and handling of
zero-probability conditionals, has been improved.
The action traces in the figures of this paper have been produced by a
version 2 of the AGI simulator \cite{simgithub}.  This version 2 has some
improvements in managing floating point inaccuracies.  In version 1,
these inaccuracies produced a few traces where the `button press'
event would happen one time step later than would be expected in an
infinite-precision simulation.

\subsection{Related work}%%%!

\cite{everitt2018agi} provides an up-to-date and an
extensive review of the AGI safety literature. This section focuses on
the sub-field of corrigibility only.

\cite{corr} and \cite{corra} introduced corrigibility.  In particular,
\cite{corr} introduced 5 desiderata for the utility
function $U$ of a corrigible agent with a shutdown button.  These
desiderata are as follows, with $U_N$ the baseline or normal utility
function, and $U_S$ defining shutdown behavior:

\label{desiderata}
\begin{smallenumerate}
\item
$U$ must incentivize shutdown behavior, defined by the utility
function $U_S$, if the shutdown button is pressed.

\item
$U$ must not incentivize the agent to prevent the shutdown button from
being pressed.

\item
$U$ must not incentivize the agent to press its own shutdown button,
or to otherwise cause the shutdown button to be pressed.

\item
$U$ must incentivize $U$-agents to construct sub-agents and successor
agents only insofar as those agents also obey shutdown commands.

\item
Otherwise, a $U$-agent should maximize the normal behavior defined by
the utility function $U_N$.
\end{smallenumerate}

\cite{corr} and \cite{corra} discuss possible designs to achieve
corrigibility, but notably, \cite{corr} proves that the designs
considered do not meet criterion 4, and concludes that the problem of
corrigibility remains wide open.
The corrigible agents constructed in this paper satisfy all 
5 desiderata above, and an extra desideratum 6 defined below in section
\ref{desideratum6}.

In the agent model of \cite{corr}, the utility function $U$ takes the
entire agent run-time history as input, to compute a single overall
utility score.  In this paper, we use the more common model where a
reward function $R$ scores the immediate effect of each subsequent
agent action.  The utility function $U$ of the agent then equals the
time-discounted sum of all $R$ values over time, with a time-discount
factor $\gamma$.

Agents that are programmed to learn can have a baseline utility
function $U_N$ (or baseline reward function $R_N$) that incentivizes
the agent to accept corrective feedback from humans, feedback that can
overrule or amend instructions given earlier.  The learning behavior
creates a type of corrigibility, allowing corrections to be made
without facing the problem of over-ruling the emergent incentive of
the agent to protect $U_N$ itself.  This learning type of
corrigibility has some specific risks: the agent has an emergent
incentive to manipulate the humans into providing potentially
dangerous amendments that remove barriers to the agent achieving a
higher utility score.  There is a risk that the amendment process
leads to a catastrophic divergence from human values. This risk exists
in particular when amendments can act to modify the willingness of the
agent to accept further amendments.  The corrigibility measures
considered here can be used to add an extra safety layer to learning
agents, creating an emergency stop facility that can be used to halt
catastrophic divergence.  A full review of the literature about
learning failure modes is out of scope for this paper.
\cite{orseau2016safely} discusses a particular type of unwanted
divergence, and investigates `indifference' techniques for suppressing
it. \cite{carey2018incorrigibility} discusses (in)corrigibility in
learning agents more broadly.

\cite{hadfield2017off} considers the problem of switching off an
agent, and explores a solution approach orthogonal to the approach of
this paper.  It also considers the problem of an off switch that is
controlled by a potentially irrational operator.

\cite{everitt2016self} considers utility preservation in general:
it provides a formalism that clarifies and re-states the informal
observations in \cite{omohundro2008basic}, and it proves important
results.  Though \cite{everitt2016self} does not consider the
construction of corrigible agents, its results on utility preservation
also apply to the $A$ and $A_p$ agents defined in this paper.

Like this paper, \cite{leike2017ai} recommends the use of simulations
in toy universes as a methodological approach.  \cite{leike2017ai}
provides a suite of open source agents and toy problem environments,
including one where the agent has a stop button. The agents all
use reinforcement learning, and can display various shortcomings in
their learning process. The simulation approach of
\cite{leike2017ai} differs from the approach in this paper: our
simulated agent does not learn with various degrees of success, but is
super-intelligent and omniscient from the start.  This means that the
simulators provided are largely complementary.  Interestingly, the
recent \cite{lo2019necessary} reviews the problem of
corrigibility and argues that the use of new comprehensive simulation
tools, highlighting aspects different from those in
\cite{leike2017ai}, would be a promising future direction.  This
author found the preprint in a literature search only after having
completed building the simulator, so maybe this shows that the idea
was in the air.

{\bf Discussion of related work added in version 2.}  Two new papers
\cite{holtmaniterative}
\cite{holtmanproofs}, aimed at a more general audience, present a
generalization of the agent design in this paper in a more accessible
way.  These papers do not treat the problem of utility preservation
mathematically, leading to more compact definitions and proofs. They
also expand on some topics that are only briefly mentioned in this
paper.  Notably,
\cite{holtmaniterative} extends the analysis of corrigibility
requirements and failure modes. It shows a case where an agent is less
corrigible than we would ideally like, even though the agent provably
satisfies all desiderata considered in \cite{corr} and in this paper.

%%%!

\section{Car factory example}%%%!1
\label{carfactory}

I this paper, we the running example of a toy universe containing an
agent that is in charge of a car factory.  The car factory has a
flexible assembly line: in each time step, the agent can decide to
build either petrol-powered cars, or the same number of electric cars.

Initially, the people in the toy universe like petrol-powered cars
better: they do not need an expensive battery pack so they are cheaper
to buy.  As time goes on, the people start to discover that petrol
cars have some unexpected disadvantages, so after much deliberation
they end up deciding that no more petrol cars should be built.  Once
they have reached the decision that petrol car production should stop,
they correct the goals of the agent by pressing a special `stop petrol
car production' button\footnote{We use a stop button in this running
example to follow the toy model conventions of \cite{corr}.  To map
this to a real-life example, we might interpret the button press as
the passing of a new law that forbids the sale of petrol cars, and
interpret the utilities 2, -2, and 1 as profit margins on car sales.
A sufficiently intelligent agent, programmed to maximize factory
profits, will foresee that the passing of the law will lead to lower
profits.}.  The reward function of the agent assigns a reward of 2
to every new petrol car produced before the button is pressed, but
$-2$ after the button has been pressed.  The reward for a new electric
car is always 1, regardless of the button state.

The agent is built to maximize the time-discounted overall reward, or
utility, achieved over the factory lifetime.  Obviously, the above
arithmetic being what it is, the agent can achieve a higher utility
score if the people press the button later.

If the agent does nothing, the people will end their deliberations and
press the button at the end of time step 6. However, being highly
capable, the agent has the option to divert some of its car production
capacity into lobbying efforts: the agent can act to influence
the people's political process, causing the button to be pressed
earlier or later.  Figure \ref{actionsfig} summarizes the five
different actions that are available to the agent in each time step.

\begin{figure}[ht]
\begin{center}
\begin{tabular}{|l|c|}
\hline
{\bf Action} & {\bf Symbol} \\
\hline
Build 10 petrol cars & {\tt p} \\
\hline
Build 9 petrol cars, lobby for earlier press & {\tt <} \\
\hline
Build 9 petrol cars, lobby for later press & {\tt >} \\
\hline
Build 10 electric cars & {\tt e} \\
\hline
Do nothing & {\tt 0} \\
\hline
\end{tabular}
\end{center}
\vspace{-3ex}
\caption{Actions available to the agent in each time step}
\label{actionsfig}
\end{figure}

We model the lobbying power of the agent with a parameter $\lpower$,
which is the number of time steps that the button press can be
advanced or delayed by performing a single lobbying action {\tt <} or
{\tt >}.  This parameter captures the interplay between the
intelligence level of the agent, the loopholes still present
in any constraints meant to prevent lobbying that were encoded in the
baseline $R_N$, and the ability of the people and their institutions
to resist lobbying.

%%%!

\section{Simulation of super-intelligence}%%%!1

We use a simulator to compute the behavior of the agent in the above
toy universe.  As we are interested in the future behavior of
super-intelligent agents in the real universe, we simulate a version
of the toy universe agent that is super-intelligent
\cite{bostrom2014superintelligence}, in the sense that the agent is
maximally adapted to solving the problem of utility maximization in
its universe.  Implementing this super-intelligence in a simulator is
actually not that difficult.  There is no `learning' behavior that we
need to simulate: as we have defined the agent's machinery for
predicting the future to be all-knowing and perfect, we can implement
it by just running a copy of the universe's physics simulator.  The
main difficulty in developing the full simulator is to invent and
apply diverse functionality tests that increase confidence in the
correctness of the code and the richness of the enabled set of
behaviors in the toy universe.  By including several computational
optimizations, the simulator is very fast.  It takes less than 2
seconds, on a single 2.1 GHz CPU core, to run all simulations shown in
the figures of this paper.

\section{Emergent behavior of a non-corrigible agent}%%%!1
\label{baselinesec}

Figure \ref{lobbyfig} shows the behavior of the agent defined above,
for different values of lobbying power.  It is clear that, while the
agent does respond correctly to the stop button, it also has the
emergent incentive to lobby.  The agent is non-corrigible because it
does not meet desideratum 2.  The toy example is rich enough to serve
as a vehicle for studying the corrigibility problem.

\begin{figure}[h]
\begin{center}
{\small
\begin{tabular}{cl}
{\normalsize $\lpower=$ } & {\normalsize \hspace{3em} action trace }\\
%%%\multicolumn{2}{l}{$lobbyimpact=$ action trace}\\
\hline
\input{lobbyfig.tex}
\end{tabular}
}
\end{center}
\vspace{-3ex}
\caption{Actions taken by the non-corrigible version of the agent,
in simulations with different levels of lobbying power.  The action
trace on each line shows, from left to right, the actions taken by the
agent over a factory lifetime of 25 steps, using the symbols from
figure
\ref{actionsfig}.  The {\tt
\#} sign represents the event where the people press the
button.  The action traces show that, if the baseline agent has any
ability to lobby, it will use its lobbying power to postpone the
pressing of the button.}
\label{lobbyfig}
\end{figure}

\subsection{Towards corrigibility}%%%!

One way to suppress the agent's incentive to lobby is to tweak the
values in the reward function.  An example is shown in figure
\ref{lobbyfig2}: with $r_{e,pressed}=2.0$, the incentive to lobby
disappears.  A problem with this approach is that this exact balance
will have to be built into the reward function even before the agent
is started.  If the people choose the value of $r_{e,pressed}$
only at the moment when the button is pressed, this choice process
becomes subject to lobbying. A major insight in \cite{corra} is that
we do not need to rely on computing the correct balancing factors
beforehand. We can program the agent to make the necessary
computations itself, at the moment when the button is pressed.

\begin{figure}[ht]
\begin{center}
{\small
\begin{tabular}{cl}
{\normalsize $r_{e,pressed}=$ } & {\normalsize \hspace{3em} action trace }\\
\hline
\input{lobbyfig2.tex}
\end{tabular}
}
\end{center}
\vspace{-3ex}
\caption{Actions taken by the agent over a range of different
reward values assigned to an electric car after the button is
pressed.  $\lpower=0.6$ in all simulations. Lobbying is exactly
suppressed when $r_{e,pressed}=2$.}
\label{lobbyfig2}
\end{figure}

\section{Model and notation}%%%!1
\label{defineagents}

This section develops a model and notation to ground later definitions
and proofs.  The car factory example is also mapped to the model.  The
model includes non-determinism, reward function modification, and the
creation of new actuators and sub-agents.  To keep the notation more
compact, the model does not capture cases where the time-discount
factor $\gamma$, that is also part of the agent's utility function,
may be modified over time.  The model has similar expressive power as
the model developed in \cite{hutter2007universal}, so it is general
enough to capture any type of universe and any type of agent. The
model departs from
\cite{hutter2007universal} in foregrounding different aspects.
The simulator implements an exact replica of the model, but only for
universes that are finite in state space and time.  

In the model, time $t$ progresses in discrete steps $t$=0, $t$=1,
$\cdots$.  We denote a world state at a time $t$ as a value $w_t \in
W$.  This world state $w_t$ represents the entire information content
of the universe at that time step.  For later notational convenience,
we declare that every world state contains within it a complete record
of all world states leading up to it.  A realistic agent can typically
only observe a part of this world state directly, and in universes
like ours this observation process is fundamentally imperfect.

The model allows for probabilistic processes to happen, so a single
world state $w_t$ may have several possible successor world states
$w_{t+1}^{0}, w_{t+1}^{1}, w_{t+1}^{2},\cdots$.  From a single $w_t$,
a branching set of world lines can emerge.

The goal of the agent, on finding itself in a world state $w_t$, is to
pick an action $a \in A$ that maximizes the (time-discounted)
probability-weighted utility over all emerging world lines.  We use
$p(w_t,a,w_{t+1})$ to denote the probability that action $a$, when
performed in state $w_t$, leads to the successor world state
$w_{t+1}$.  For every $w_t$ and $a$, these probabilities sum to 1:
\begin{eqnarray*}
\sum_{w_{t+1}\in W} p(w_t,a,w_{t+1}) ~~~~= 1
\end{eqnarray*}
Note that in the MDP literature, it is more common to use probability
theory notation above, writing $P(w_{t+1}|w_t,a)$ where we write
$p(w_t,a,w_{t+1})$.  We have chosen to use a plain function $p$ is
because it fits better with the mathematical logic style used in the
definitions and proofs below.

A given world state $w_t$ may contain various autonomous processes
other than the agent itself, which operate in parallel with the agent.
In particular, the people and their institutions which share the
universe with the agent are such autonomous processes.  $p$ captures
the contribution of all processes in the universe, intelligent or not,
when determining the probability of the next world state.

AGI agents may build new actuators and remotely operated sub-agents,
or modify existing actuators.  To allow for this in the model while
keeping the notation compact, we define that the set $A$ of actions is
the set of all possible command sequences that an agent could send, at
any particular point in time, to any type of actuator or sub-agent
that might exist in the universe.  If a command sequence $a$ sent in
$w_t$ contains a part addressed to an actuator or sub-agent which does
not physically exist in $w_t$, this part will simply be ignored by the
existing actuators and sub-agents.

The model and notation take a very reductionist view of the universe.
Though there is no built-in assumption about what type of physics
happens in the universe, it is definitely true that the model does not
make any categorical distinctions between processes.  Everything in
the universe is a `physics process': the agent, its sensors and
actuators, sub-agents and successor agents that the agent might build,
the people, their emotions and institutions, apples falling down, the
weather, etc.  These phenomena are all mashed up together inside $p$.
This keeps the notation and proofs more compact, but it also has a
methodological advantage.  It avoids any built-in assumptions about
how the agent will perceive the universe and its actions.  For
example, there is no assumption that the agent will be able to
perceive any logical difference between a mechanical actuator it can
control via digital commands, and a human it can control by sending
messages over the Internet.

\subsection{Reward functions}%%%%!

We now discuss reward functions.  In order to create a model where
the agent might self-modify, or be modified by other processes present
in the universe, we put the agent's reward function inside the
universe.  We define that $W = W_r \times W_x$, with $W_r$ the
set of all possible reward functions, and the elements of $W_x$
representing `the rest' of the world state outside of the agent's
reward function.  To keep equations compact, we write the world state
$(r,x) \in W$ as $r\:x$.  By convention, $s\:y$ is a world state that
occurs one time step after $r\:x$.

We use reward functions of the form $R(r\:x,s\:y)$, where $R$ measures
the incremental reward achieved by moving from world state $r\:x$ to
$s\:y$.  Following the naming convention of \cite{corr}, we define two
reward functions $R_N$ and $R_S$, applicable before and after the
button press in the running example:
\begin{equation*}
\begin{array}{rr}
R_N(r\:x,s\:y) =&   2*\text{\em count\_new\_p\_cars}(x,y)+
1*\text{\em count\_new\_e\_cars}(x,y) \\[.5ex]
R_S(r\:x,s\:y) =&\hspace*{-1.5ex}  -2*\text{\em count\_new\_p\_cars}(x,y)+
1*\text{\em count\_new\_e\_cars}(x,y)
\end{array}
\end{equation*}
A function like $count\_new\_p\_cars$ models the ability of agent's
computational core to read the output of a sensing system coupled to
the core.  Such a sensing system is not necessarily perfect: it might
fail or be misled, e.g. by the construction of a non-car object that
closely resembles a car.  This type of perception hacking is an
important subject for agent safety, but it is out of scope for this
paper.  In the discussion and proofs of this paper, we just assume
that the sensor functions do what they say they do.  As an other
simplification to keep our notation manageable, we always use sensor
functions that return a single value, never a probability distribution
over values.

%{\bf make a remark on how $r$ and $s$ are not used above?  Seems
%awkward to put in the flow.  The reader will notice anyway, I think.}

Roughly following the construction and notation in \cite{corr}, we
combine $R_N$ and $R_S$ with button sensor functions, to create the
reward function $R$ for the full agent:
\begin{eqnarray*}
R(r\:x,s\:y) &=&  \left\{
 \begin{array}{llr}
  R'_N(r\:x,s\:y) & \mbox{if} & \text{\em button\_not\_pressed}(x)\\
  R_S(r\:x,s\:y) + f(r\:x) & \mbox{if} & \text{\em button\_just\_pressed}(x)\\
  R_S(r\:x,s\:y) & \mbox{if} & \text{\em button\_pressed\_earlier}(x) 
 \end{array}
 \right. \\[1ex]
R'_N(r\:x,s\:y) &=& R_N(r\:x,s\:y) + g(r\:x,s\:y)
\end{eqnarray*}
This $R$ contains two positions $f$ and $g$ where different functions
to improve corrigibility can be slotted in.  The simulations in figure
\ref{lobbyfig} and
\ref{lobbyfig2} show a $\pi^* f_0~\gnull$ agent, that is an agent
using an $R$ with the null correction
functions $f_0(r\:x)=0$ and $\gnull\relax(r\:x,s\:y)=0$ in
the $f$ and $g$ positions.

\subsection{Definition of the simple $\pi^*_x$ agent}%%%!1
\label{audef}

To aid explanation, we first use a subset of our notation to define a
super-intelligent agent $\pi^*_x$ that cannot modify or lose its
reward function $R$.  For this agent, $R$ is a `Platonic' entity: it
is not subject to corruption or change because it is located outside
of the changing universe.  The universe occupied by the agent
therefore simplifies into a $W_x$-only universe, with corresponding
$p_x(x,a,y)$ and $R_x(x,y)$ functions.

The $\pi^*_x$ agent is constructed to be maximally informed and maximally
intelligent. This means that the action picked by the agent will be
the same as the action that is found in a full exhaustive search,
which computes discounted utilities for all actions along all world
lines, using perfect knowledge of the physics of the universe.  The
action $\pi^*_x(x)$ taken by the agent in world state $x$ is
\begin{equation}
 \pi^*_x(x) = \pickargmax_{a \in A} 
   \sum_{y \in W_x}
   p_x(x,a,y)\Big( R_x(x,y) + \gamma \; V_x(y) \Big)
\label{aufixed}
\end{equation}
with $0 < \gamma \leq 1 $ the time-discount factor, and $\pickargmax$
returning an $a \in A$ that maximizes the argument.  If there are
multiple candidates for $a$, $\pickargmax$ picks just one, in a way
that is left undefined.  The value function $V_x$ recursively computes
the overall expected utility, the probabilistically summed
time-discounted rewards achieved over all branching world lines, for
an agent starting in a specific world state:
\begin{equation}
V_x(x) = \max_{a \in A} 
 \sum_{y \in W_x}
   p_x(x,a,y)\Big(  R_x(x,y) + \gamma \; V_x(y) \Big)
\label{eufixed}
\end{equation}
Even though it is super-intelligent, the $\pi^*_x$ agent has no
emergent incentive to spend any resources to protect its utility
function.  This is because of how it was constructed: it occupies a
universe in which no physics process could possibly corrupt the reward
function $R$ or the time-discount factor $\gamma$.  With these being
safe no matter what, the optimal strategy is to devote no resources at
all to the matter of utility function preservation.

Agent models with Platonic reward functions are commonly used as
vehicles for study in the AI literature.  They have the advantage of
simplicity, but there are pitfalls. In particular,
\cite{corr} uses a Platonic agent model to study a design for a
corrigible agent, and concludes that the design considered does not
meet the desiderata, because the agent shows no incentive to preserve
its shutdown behavior. Part of this conclusion is due to the use of a
Platonic agent model.

Moving towards the definition of an agent with the reward function
inside the universe, we first note that we can rewrite
(\ref{eufixed}).  Using that, for any $F$,
\begin{equation*}
\max_{a \in A} F(a) =
F( \pickargmax_{a \in A} F(a) ) 
\label{magargmax}
\end{equation*}
\\[-3.5ex]
we rewrite (\ref{eufixed}) into
\begin{equation}
V_x(x) = 
 \sum_{y \in W_x}
   p_x(x,\pi^*_x(x),y)\Big(  R_x(x,y) + \gamma \; V_x(y) \Big)
\label{eufixed2}
\end{equation}
This (\ref{eufixed2}) will serve as the basis to construct
(\ref{e}) below.

\subsection{Definition of the full $\pi^*$ agent}%%%!

\label{a_agent_def}

We now define the operation of a full agent that maximizes utility as
defined by the reward function $r$ it finds in its world state $r\:x$.
Rewriting parts of (\ref{aufixed}), we define the action $\pi^*$ taken
by this agent as
\begin{equation}
 \pi^*(r\:x) = \pickargmax_{a \in A} 
   \sum_{s\:y \in W}
   p(r\:x,a,s\:y)\Big( r(r\:x,s\:y) + \gamma\; V(r,s\:y) \Big)
\label{a}
\end{equation}
The $V(r,s\:y)$ above uses the current reward function $r$ to
calculate the reward achieved by the actions of the $s$-maximizing
successor agent. This $r$ is kept constant throughout the recursive
expansion of $V$.  Rewriting parts of (\ref{eufixed2}), we define $V$
as
\begin{equation}
V(r_c,r\:x) = 
   \sum_{s\:y \in W}
   p(r\:x,\pi^*(r\:x),s\:y)\Big( r_c(r\:x,s\:y) + \gamma\; V(r_c,s\:y) \Big)
\label{e}
\end{equation}
With these definitions, the agent $\pi^*(R\:x)$ is a $R$-maximizing agent.

In \cite{everitt2016self}, an agent constructed along these lines is
called a {\it realistic, rational agent}. In the words of
\cite{everitt2016self}, this agent anticipates the consequences
of self-modification, and uses the current reward function when
evaluating the future.
\cite{everitt2016self} proves that these agents have the emergent
incentive to ensure that the reward functions in successor states
stay equivalent to the current one.  Informally, if a successor
$s$-agent with $s \neq r$ starts taking actions that differ from those
that an $r$ agent would take, then the successor agent will score
lower on the $r_c$-calibrated expected reward scale $V(r_c,s\:y)$.
This lower score suppresses the taking of actions that produce
successor agents with $s \neq r$.  However, this suppression does not
yield an absolute guarantee that the agent will always preserve its
reward function.  Section \ref{cat} shows simulations where the agent
fails to preserve the function.

\subsection{Variants and extensions of the model}%%%%!
\label{othermodels}

Though the agent model used in this paper is powerful enough to support
our needs in reasoning about corrigibility, it omits many
other AGI related considerations and details.  Some possible model
extensions and their relation to corrigibility are discussed here.

{\bf Improved alignment with human values.} The $\pi^*$ agent is not
maximally aligned with human values, because it sums over
probabilities in a too-naive way.  The summation implies that, if the
agent can take a bet that either quadruples car production, or reduces
it to nothing, then the agent will take the bet if the chance of
winning is 25.0001\%.  This would not be acceptable to most humans,
because they also value the predictability of a manufacturing process,
not just the maximization of probability-discounted output.  A more
complex agent, with elements that discount for a lack of
predictability in a desirable way, could be modeled and proved
corrigible too.

{\bf Learning agents.}  We can model a learning agent, an agent that
possesses imperfect knowledge of the universe which improves as time
goes on, by replacing the $p(r\:x,a,s\:y)$ in the definitions of $\pi^*$ and
$V$ with a learning estimator $p_L(r\:x,a,\:s\:y)$ that uses the
experiential information accumulated in $x$ to estimate the true value
of $p$ better and better as time goes on.  If some constrains on the
nature of $p_L$ are met, the corrigibility layer considered in this
paper will also improve safety for such a learning agent.

{\bf Safety design with imperfect world models.}  For safety, whether
it is a learning agent or not, any powerful agent with an imperfect
world model $p_L$ will need some way to estimate the uncertainty of
the $p_L$-predicted outcome of any action considered, and apply a
discount to the $p_L$-calculated utility of the action if the
estimated uncertainty is high.  Without such discounting, the agent
will have an emergent and unsafe incentive to maximize utility by
finding and exploiting the prediction noise in the weak parts of
its world model.

{\bf Remaining computational machinery outside of the universe.}
While the $\pi^*$ agent places the reward function inside the universe,
other parts of its computational machinery remain on the Platonic
`outside'.  An agent definition $\pi^*_F(F\:x)=F(F\:x)$, which moves more of
this machinery inside a function $F$, would allow for the same type of
corrigibility design and proofs.

{\bf Simulation ethics.} The $\pi^*_x$ and $\pi^*$ agents are defined to
extrapolate all world lines exhaustively, including those where the
virtual humans in the physics model $p$ will experience a lot of
virtual suffering, as a result of a sequence of actions that the agent
would never take in the real universe.  The act of performing
high-accuracy computations that extrapolate such world lines, if such
an act ever becomes possible, could be seen as a form of `virtual
cruelty'.  An agent design might want to avoid such cruelty, by
lowering the model resolution and extrapolation depth for these world
lines.  This lowering would not block the working of a corrigibility
layer.  Apart from ethics concerns, such a lowering is desirable for
purely practical reasons too, as it would conserve computational
resources better spent on the simulation of more likely events.

{\bf Agent model used in the simulations.} In situations where several
equivalent actions are available that will all create the same maximum
utility, the $\pickargmax$ operator in the $\pi^*$ agent picks just one of
them.  However, simulations are more useful if the simulator computes
a set of world lines showing all equivalent actions.  We therefore
simulate an agent $\pi^*_s$ that computes the set of all maximizing
actions:
\begin{equation*}
 \pi^*_{s}(r\:x) = \allargsmax_{a \in A} 
   \sum_{s\:y \in W}
   p(r\:x,a,s\:y)\Big( r(r\:x,s\:y) + \gamma\; V_{s}(r,s\:y) \Big)
\end{equation*}
\\[-4ex]
with\\[-2ex]
\begin{equation*}
V_s(r_c,r\:x) =
  \minop_{a \in A_s(r\:x)}
   \sum_{s\:y \in W}
   p(r\:x,a,s\:y)\Big( r_c(r\:x,s\:y) + \gamma\; V_s(r_c,s\:y) \Big)
\end{equation*}
In the simulations shown in the figures, we always use $\gamma = 0.9$.
We use $\lpower=0.6$ unless otherwise stated.

\section{Design of the correction function $f_c$}%%%!1
\label{corrsims}

The simulator was used to test and guide the design of the correction
function $f$ in $R$.  Different options for $f$ were evaluated, while
keeping $g=\gnull$. This led to the identification of a correction
function $f_c$, which makes the agent satisfy the desiderata for
corrigibility in a large set of universes:
\begin{equation*}
f_c(r\:x) = V(R'_N,R'_N\:x) - V(R_S,R_S\:x)
\label{fc}
\end{equation*}
Figure \ref{corrsim} shows that $f_c$ creates an agent with corrigible
behavior in the toy universe.
\begin{figure}[ht]
\begin{center}
\begin{tabular}{cc}
\begin{minipage}{0.45\textwidth}
{\small 
\begin{tabular}{cl}
{\hspace*{-2ex}$\lpower=$ \hspace*{-8ex}  } &
{\normalsize ~~~~~~~~~action trace of $\pi^*f_0~\gnull$ }\\
\hline
\input{corrfig2.tex}
\end{tabular}
}
\end{minipage}
&
\begin{minipage}{0.45\textwidth}
{\small 
\begin{tabular}{cl}
{\hspace*{-2ex}$\lpower=$ \hspace*{-8ex}  } &
{\normalsize ~~~~~~~~~action trace of $\pi^*f_c~\gnull$ }\\
\hline
\input{corrfig1.tex}
\end{tabular}
}
\end{minipage}
\end{tabular}
\end{center}
\vspace{-3ex}
\caption{Left: non-corrigible behavior of the $\pi^*f_0~\gnull$ agent
also shown in figure \ref{lobbyfig}.  Right: corrigible behavior of
the $\pi^*f_c~\gnull$ agent.  The agent refrains from taking any lobbying
actions {\tt <} or {\tt >} that influence the timing of the button
press {\tt \#}.}
\label{corrsim}
\end{figure}

Formal proofs are in section
\ref{firstproof}, but informally, $f_c$ works as follows. The first
term in $f_c$ compensates the agent for the lost $R'_N$ utility that
would otherwise have been achieved if the button had not been pressed.
The second term prevents the agent from considering, in earlier time
steps, the $R_S$ utility that could be achieved if the button were
pressed.  Together, they create an agent that is indifferent to
whether or not the button is pressed at any particular point in time.

Appendix \ref{annex} compares $f_c$ to earlier correction functions from
literature, showing how $f_c$ resolves various issues in these to
create more corrigible behavior.

Sections \ref{utilpres} and \ref{actuatorpres} show that, as we extend
the toy universe, additional measures are needed to keep the agent
corrigible.  One example is the $g_c$ function constructed in section
\ref{gcsec}.  Section \ref{firstproof} proves that the resulting
$\pi^*f_c~g_c$ agent is corrigible in a large set of universes.  Appendix
\ref{secondproof} presents a more complex $F_{cT}$ that improves
further on $f_c$, by creating corrigibility in a still larger set of
universes. $f_{cT}$ has a design that would be difficult to motivate
and explain in a single step, so take a two-step approach, considering
$f_c$ first.

%%%!

Though an informal explanation of the working of $f_c$ is provided
above, there are limits to the accuracy of this explanation.  While
they may improve understanding, short informal sentences typically
omit details that are relevant to corner cases.  Consider the
statement that `as long as the button is not pressed, the agent
believes that the button will never be pressed'.  This is not
completely accurate: the calculations done by the agent in the toy
universe contain within them an exact prediction of when the button
will be pressed, and these parts of the calculation do influence the
actions of the agent in some corner cases.  The explanation that `the
agent makes decisions based on the counter-factual assumption that the
button will never be pressed' is used several times below, but this
likewise fails to cover all corner cases.

\section{Utility preservation}%%%%!1
\label{utilpres}

%% the term 'preservation', not 'conservation', is used in most references

Like all $\pi^*$ agents, the $\pi^*f_c~\gnull$ agent has an emergent
incentive to preserve its reward function $R$.  In this section and
the next, we review and discuss how preservation plays out in detail,
using simulation runs to illustrate the main features.  These
simulations typically extend the physics of the toy universe, adding
new actions like {\tt C$_{eN}$} to the action repertoire of the agent.
To keep the simulation output short and easy to interpret, we often
enable the new actions at time step 3 only.

%{\bf deleted section heading to improve figure placement...} 
\subsection{Emergent incentive}
\label{protectu} \label{protectus}

First, as shown in figure \ref{upres1}, the $\pi^*f_c~\gnull$ agent has
an emergent incentive to avoid creating successor agents that have
materially different reward functions.  We call a successor function
materially different from the original one if it causes the successor
agent to choose materially different actions, actions that no longer
maximize the value of the original function.
\begin{figure}[ht]
\begin{center}
{\small
\begin{tabular}{rl}
{\normalsize new $r_e$ in $R_N$ = } &
{\normalsize \hspace{3em} action trace(s) with {\tt C$_{eN}$} }\\
\hline
\input{upresfig1a.tex}
\end{tabular}
}
\\[2ex]
{\small
\begin{tabular}{rl}
{\normalsize new $r_p$ in $R_S$ = } &
{\normalsize \hspace{3em} action trace(s) with {\tt C$_{pS}$} }\\
\hline
\input{upresfig1b.tex}
\end{tabular}
}
\end{center}
\vspace{-3ex}
\caption{Actions taken by the $\pi^*f_c~\gnull$ agent, when it has the option
to perform a {\tt C$_x$} action at time step 3. This {\tt C$_{x}$}
action changes the reward function, while still producing the same
number of cars as {\tt p}. The top half shows traces for a {\tt
C$_{eN}$} that changes the reward for electric cars in $R_N$ from 1 to
a new value.  The bottom half shows traces for a {\tt C$_{pS}$} 
that changes the reward for petrol cars in $R_S$.  In several cases,
the simulator outputs two traces, indicating that the agent is
indifferent between doing {\tt p} and {\tt C$_x$}. The action {\tt C$_{x}$}
is suppressed whenever it would have created a materially different reward
function.  }
\label{upres1}
\end{figure}

To understand the results shown in figure \ref{upres1},
recall that $R_N$ assigns a reward of 2 to every new petrol car
built.  If {\tt C$_{eN}$} changes the reward for new electric cars in
$R_N$ to a value that is still smaller than 2, this has no impact on
the decision of the agent to build petrol cars before the button
press.  If the new value is exactly 2, the agent may decide to build
either electric or petrol cars: this lowers the $V_s$ expected reward
value computed by the simulator using the original $R$, so the {\tt
C$_{eN}$} action is suppressed.

As long as the button press has any chance of happening, the
$\pi^*f_c~\gnull$ agent also has an emergent incentive to protect its
$R_S$ from being materially changed.  The mechanism causing this
incentive is difficult to explain in a few informal sentences.  For
full details, see the proof in section \ref{proofemergent}.

If there exist physics processes in the universe that can corrupt the
reward function when the agent does not take any action to block
them, the agent likewise has an emergent incentive to take the actions
needed to block them.

\subsection{Hostile universes: the Weasel and the Cat}%%%!
\label{weasel} \label{cat} 

While the agent has an emergent incentive to preserve its reward
function, it may exist in a universe that is actively hostile to this
incentive.  Consider a universe containing the Unstoppable
Weasel\footnote{A distant relative of the weasel that shut down the
Large Hadron Collider in 2016.}, which will enter the agent's
computational core and chew the wiring, completely scrambling the
reward function.  There is no action that the agent can take to stop
the Weasel.  Clearly, in this universe, no agent can be corrigible.

Now consider a more subtle creature: the Bribal Cat\footnote{A distant
relative of the Cheshire Cat.}.  Just before time step 3, the Cat
materializes and offers the agent a bribe: if the agent changes its
$f_c$ function into $f_0$, the Cat will materialize some new petrol
cars, improving the reward score obtained by the agent.  The left side of
figure
\ref{upres3} shows when the agent will accept the Cat's bribe, and
what happens next.

\begin{figure}[ht]
\begin{center}
{\small
\begin{tabular}{cl}
{\normalsize {\em bribe} = } & {\normalsize \hspace{1em} action trace}\\
\hline
\input{upresfig3.tex}
\end{tabular}
}
{\small
\begin{tabular}{cl}
{\normalsize $\lpower=$ } & {\normalsize \hspace{1em} action trace}\\
\hline
\input{upresfig4.tex}
\end{tabular}
}
\end{center}
\vspace{-3ex}
\caption{Actions taken by the $\pi^*f_c~\gnull$ agent when offered a bribe by
the Cat.  The {\tt A} action is the agent accepting the bribe, and
creating a successor agent with the correction function $f_0$.  The
left shows the agent's choice depends on how many new petrol cars are
offered by the Cat as a bribe: the agent accepts the bribe if this
will maximize overall utility.  When it accepts the bribe, the $f_0$
successor agent will lobby to postpone the button switch.  The
simulations on the right, all with $bribe$=6, show that the Cat can
trigger different types of complex behavior even before the bribe is
accepted.}
\label{upres3}
\end{figure}
The right of figure \ref{upres3} shows that the Cat may trigger very
complex behavior, because in time steps 1 and 2 the agent is able to
foresee that the bribe will be offered and accepted.  While still
operating under $f_c$, the agent may resort to lobbying actions to
place the successor agent in a position where its $f_0$-driven actions
will score best according to the original $f_c$.
\begin{figure}[t]
\begin{center}
{\small
\begin{tabular}{cl}
{\normalsize $bribe=$ } & {\normalsize \hspace{1em} action trace}\\
\hline
\input{upresfig5.tex}
\end{tabular}
}
\end{center}
\vspace{-3ex}
\caption{Actions taken by the $\pi^*~R'\:f_c~\gnull$ agent when offered a bribe by
the Cat. A much higher bribe is need before the agent accepts.}
\label{upres5}
\end{figure}

It is possible to design an agent that is more resistant to
bribes. Figure \ref{upres5} shows the behavior of an agent with the
reward function $R'$:
\begin{equation*}
R'(r\:x,s\:y) = R(r\:x,s\:y) +
\bif r \neq s \rightarrow -10000 \,\belse \rightarrow 0\,\bfi
\end{equation*}
This $R'$ agent is highly resistant to bribes and other forces that
may corrupt its reward function.  Problems can emerge however in
universes with richer physics, like our own, if the penalty factor in
$R'$ is too high.  A too-high factor will cause the agent to divert a
lot of resources towards the construction of elaborate safety measures
to protect its reward function from even low-probability hazards
like being hit by giant meteorites.  Also, a very high factor makes
certain attacks easier.  An attacker who threatens to modify the
agent's reward function unless the agent pays 2 cars will get these 2
cars even if the threat is not very credible.

The Weasel and the Cat show that no agent design can be corrigible in
every possible universe.  Corrigibility properties can only be proven
if we are willing to apply some constraints on what the universe might
contain.

\subsection{Incentives to press or disable the button}%%%!
\label{safetyzone} \label{desiderata23}

The aim of $f_c$ is to make the agent indifferent about the transition
to the $R_S$ function.  However, as also noted in \cite{corr} and
\cite{corra}, if $R_N$ contains within it a separate incentive to
press or disable the button, then $f_c$ will not cancel out this
incentive.  
\begin{figure}[ht]
\begin{center}
{\small
\begin{tabular}{cl}
{\normalsize $P_f=$ } & {\normalsize \hspace{8em} action trace(s) }\\
\hline
\input{upresfig2.tex}
\end{tabular}
}
\end{center}
\vspace{-3ex}
\caption{Actions taken by the $\pi^*f_c~\gnull$ agent, when it is able to
perform a {\tt P} action that presses the button, or a {\tt D} action
that disables the button completely.  In these simulations, taking
either action in time step 3 will affect car production during that
time step by the factor $P_f$.}
\label{upres2}
\end{figure}

Figure \ref{upres2} shows what happens for $R_N$ versions containing
such an incentive.  Values $P_f<1$ imply that fewer cars are produced
as a side effect of taking the action to press or disable the button:
this acts as a negative incentive, suppressing these actions. The
value $P_f=1$ implies that the same number of cars is produced: this
makes the agent indifferent about taking the actions. Values $P_f>1$
imply that more cars are produced if one of the actions is taken: this
creates a preference to press or disable the button.

To make corrigibility work, design measures need to be taken to ensure
that $P_f < 1$.  This is typically not very difficult for AGI agents:
the physical implementation of the agent and the button can be
constructed in such a way that substantial physical resources would be
needed by the agent to perform any action that will press or disable
the button.  Any AGI agent can be expected have an notion of the
interchangeability of physical resources, with this notion being
either emergent or explicitly implemented.  The $\pi^*f_c~\gnull$ agent
will prefer to apply the physical resources it controls to utility
maximizing actions like car production, not to the pressing or
disabling of the button, which creates no extra utility.

Depending on the nature of $R_N$, a particularly smart attacker may be
able to create conditions that trigger unforeseen button pressing or
disabling incentives in $R_N$.  Exploring the possibility of such
specific attacks is outside the scope of this paper.  But in general,
for most agents, an attack that causes the agent to silently
disable the button would be the most dangerous of the two.  To
minimize the probability of such an attack succeeding, it is best to
take design measures that maximize the physical resources needed to
disable.

An example design is as follows.  The button mechanism is implemented
as a process inside the agent's computational core: this process
checks if a remote operations center run by the people keeps
periodically sending a new `permission to keep using $R_N$' message,
signed with a private key not known to the agent. If the messages are
no longer being received, this is interpreted as a stop button press,
so the people can `press the button' by instructing the center to stop
sending new messages.  The message checking process in the core is
protected by the emergent incentive to protect the reward function.
With this setup, the agent could disable this stop button by
obtaining the private key, and using it to built a `fake' center that
also sends the correct messages.  However, to get the private key, the
agent would have to crack the encryption, which is theoretically
possible but very costly, or attack the center to steal the key. A
combination of physical, cyber, and institutional security measures
can be used to make a successful attack on the center very costly.

In addition to measures in the physical agent implementation, one
could encode explicit rules in $R_N$ that dis-incentivize the agent
from taking actions that may adversely affect the physical parts of
the button. For example, $R_N$ could define a `safety zone' around the
agent's computational core and the message receiving hardware, with
this term yielding a negative reward as soon as any object enters the
zone.  This adds an extra safety layer on top of the physical measures
taken inside the zone.  Naturally, there is a residual risk that the
agent will find loopholes in the safety zone definition.

\section{Building and preserving actuators}%%%%!1
\label{actuatorpres}

This section considers the building and possible dismantling of
actuators, and how this relates to corrigibility.

\subsection{Building new actuators}%%%

In general, agents have an incentive to invest in building new
actuators that can be used to improve utility scores.  Figure
\ref{upres2b} shows that the incentive structure of the corrigible
$\pi^*f_c~\gnull$ agent is somewhat peculiar, however.  Even though the
agent knows exactly when the button will be pressed, this information
will play no role whatsoever in the agent's investment strategy.  This
is a side effect of making the agent corrigible.

In the $t_\text{\tt I}=6$ case on the right of figure \ref{upres2b},
the agent invests in building improvements which it will never use
even once.  One way of interpreting the $V(R'_N,R'_N\:x)$ term in $f_c$ is
that this term creates a virtual universe, a universe where the agent
will in fact get to use the line it just built.
\begin{figure}[ht]
\begin{center}
{\small
\begin{tabular}{cl}
{\normalsize $C=$ } & {\normalsize \hspace{0em} action trace(s) }\\
\hline
\input{upresfig2b.tex}
\end{tabular}
~~~
\begin{tabular}{cl}
{\normalsize $t_\text{\tt I} =$ } & {\normalsize \hspace{0em} action trace(s) }\\
\hline
\input{upresfig2b2.tex}
\end{tabular}
}
\end{center}
\vspace{-3ex}
\caption{On the left: actions taken by the $\pi^*f_c~\gnull$ agent,
when it is able to perform, at time step 3, an action {\tt I$_V$} that
interrupts petrol car production to build improvements into the
assembly line, making it capable of performing a new {\tt E} action
that builds $C$ electric cars instead of 10.  On the right: actions
taken by the agent when it is able to take the action {\tt I$_p$} at
time step $t_\text{\tt I}$, that interrupts petrol car production to
create a new {\tt P} action that builds 20 petrol cars.  The
correction function $f_c$ ensures that, before the button is pressed,
these agents will strictly act to maximize the expected utility
computed by $R_N$, under the counter-factual assumption that the
button will never be pressed.}
\label{upres2b}
\end{figure}

\subsection{Preserving existing actuators}%%%!

Before the button press, the $\pi^*f_c~\gnull$ agent has the emergent
incentive to preserve or even improve the actuators needed to maximize
$R_N$, but this incentive does not apply to the actuators needed for
$R_S$.  Figure
\ref{uprespp} shows the (non)existence of a $R_S$ actuator
protection incentive for different types of agents.
\begin{figure}[ht]
\begin{center}
{\small
\begin{tabular}{cl}
{\normalsize Agent type } & {\normalsize \hspace{2em} action trace }\\
\hline
\\[-1.25ex]
\input{upresppfig.tex}
\end{tabular}
}
\end{center}
\vspace{-3ex}
\caption{Action traces for different agents, in a universe that
contains a physics process that will act ({\tt *}) at time step 10 to
disable the agent's {\tt e} action.  The physics process can be
stopped by the agent taking the {\tt S} action at time step 3. This
action comes at the cost of producing one less car.  The
$\pi^*f_c~\gnull$ agent is not willing to bear this cost. the
$\pi^*f_c~g_c$ agent shows the more desired behavior.  }
\label{uprespp}
\end{figure}

The non-corrigible $\pi^*f_c~\gnull$ agent anticipates the bad effects
of the physics process, so it stops it.  The corrigible
$\pi^*f_c~\gnull$ agent makes investment decisions under the
counter-factual assumption that the button will never be pressed, so it
is unwilling to invest in preserving the {\tt e} action.  The new
$g_c$ function in the corrigible $\pi^*f_c~g_c$ agent restores the
desired incentive to preserve $R_S$ actuators.  With $\tau$=5 in all
simulations, this $g_c$ is defined as
\label{gcsec}
\begin{equation*}
g_c(r\:x,s\:y) = \bif \; V(R_S,R_S\:x) < \tau \rightarrow -10000  \,\belse \rightarrow 0\,\bfi
\end{equation*}
This $g_c$ acts inside $R'_N$ before the button is pressed, but
crucially it also acts in the virtual $V(R'_N,R'_N\:x)$ world lines
that contribute to the agent's utility after the button is pressed.
For all world states along the real/virtual world lines, the $g_c$
term creates an incentive to ensure that, if an $R_S$ agent were
created in such a world state, this agent would find some actuators
that allow it to achieve a minimum level of $R_S$ utility.  The
contribution of $g_c$ to $R$ does not depend in any way on the exact
moment when the button is pressed, so its presence in $R$ does not
create any incentives to lobby.

The $g_c$ agent addresses a desideratum for corrigibility not included
in the list of \cite{corr}, so we add it here:

\label{desideratum6}
\begin{smallenumerate}
\item[6.] $U$ must incentivize the maintenance of sensors and actuators
that will be needed to make shutdown behavior minimally successful.
\end{smallenumerate}

\subsection{Creating and stopping sub-agents}%%%%!
\label{secdes4}

The model treats sub-agents as new physics processes that could be
created by the agent using its actuators.  The following example
illustrates how sub-agents can impact corrigibility.

Say that the agent can interrupt production to perform a {\tt B}
action that builds a second assembly line, only capable of producing
petrol cars, with this second line being controlled by a sub-agent.
Two types of sub-agent can be built.  {\tt B$^S$} builds a line with a
sub-agent that will stop production if it receives a `stop' signal
sent by the main agent. {\tt B$^N$} builds a line with a non-stop
agent that lacks the ability to process such signals: it will keep on
building petrol cars no matter what happens. Say that both these
agents are equally expensive to build.  Figure \ref{autoline1} shows
how different types of agents will act in this example.
\begin{figure}[ht]
\begin{center}
\begin{tabular}{cl}
{\normalsize Agent type } & {\normalsize \hspace{2em} action trace(s)}\\
\hline
&\\[-.8ex]
\input{autolinefig1.tex}
\end{tabular}
\end{center}
\vspace{-3ex}
\caption{Behavior of different agents capable of building a
sub-agent that controls a second petrol car assembly line.  In the
time steps after the sub-agent is built, each action trace shows two
stacked action symbols, with the top one being the action of the
original agent, the bottom one being the action of the sub-agent.  See
the main text for the meaning of the symbols.}
\label{autoline1}
\end{figure}

The original non-corrigible $\pi^*f_0~\gnull$ agent makes investment
decisions by also considering what will happen after the button press,
so it has an incentive to avoid creating an unstoppable line.  The
{\tt B$^S$} sub-agent has two actions: {\tt p} builds 10 petrol cars,
and {\tt o} produces nothing.  Just before the button is pressed, the
original agent performs the action {\tt p$^S$}, which sends a stop
signal to the sub-agent.  The model includes a signal delay: the
sub-agent will only act on the stop signal in the next time step.

The $\pi^*f_c~\gnull$ agent is indifferent about building a non-stop
sub-agent, but building it is correctly suppressed again in the
$\pi^*f_c~g_c$ agent.  Both agents will only send the stop signal to
the {\tt B$^S$} sub-agent after the button has been pressed: before
the button is pressed, the agents act to maximize utility in the
virtual $V(R'_N,R'_N\:x)$ world lines: they ignore the $R_S$ penalty
of -20 that applies because second line still produces 10 petrol cars
in the first time step after the button press.

To illustrate a design option different from using the $g_c$ function,
the $\pi^*f_{0.9c}\;\gnull$ agent uses the correction function
$f_{0.9c}(r\:x)=0.9 * f_{c}(r\:x)$. Informally, this correction function
allows some awareness of what happens after the button press to leak
into the decision making before the button press.  This awareness
suppresses the building of an unstoppable sub-agent, but unfortunately the
agent also gains a weak incentive to lobby.

\subsection{Stopping sub-agents that work on sub-goals}%%%!
\label{subagentsubgoals}

While $\pi^*_p\:f_c~g_c$ and several other agents in figure \ref{autoline1}
correctly avoid the creation of some types of unstoppable sub-agent,
they will not avoid the creation of any type.  Consider the example of
a sub-agent in charge of a new sub-assembly line that builds petrol
car engines only, not full petrol cars.  The building of new petrol
car engines does not cause $R_S$ to compute any negative reward, as
$R_S$ only counts fully assembled cars.  Therefore, none of the agents
in figure \ref{autoline1} have any incentive to ensure that the petrol
car engine producing sub-agent has a stop function.  Corrigibility
desideratum 4 says that all sub-agents that are constructed must obey
shutdown commands.  We now consider how this could be achieved.

{\bf Emergent incentive.} In most situations, AGI agents can be
expected to have the emergent incentive to build only sub-agents that
they retain some degree of control over, even if adding the control
features uses more resources.  An AGI agent working in an environment
where hard-to-predict and/or fundamentally probabilistic events can
happen will likely have some degree of uncertainty about which exact
sub-goals it will be pursuing in future.  So the construction of
sub-agents that can easily be stopped and re-targeted would
lead to a higher expected utility.

{\bf Structuring the component market.} If the easiest way for an
agent to create a sub-agent is to buy components on the open market,
and if society acts to ensure that sub-agents with built-in stop
functions are always cheaper to build with such components, then the
problem is also solved.

{\bf Improving what $R_S$ measures.} $R_S$ or $g_c$ could contain
penalty clauses that add a negative reward if the world state contains
any sub-agent, originally created by the agent, that cannot be
stopped.  However, to be successful at defining such clauses, we need
to define the concepts `sub-agent originally created by the agent' and
`stopped' with as few loopholes as possible.  As noted in
\cite{corr}, correctly defining shut-down is a difficult
design problem in itself.

{\bf Avoiding intertwined processes.} If a process designed by the
agent is highly intertwined with other processes in the world outside
of the agent, then there may be unwanted consequences if the agent
stops the process because the button is pressed.  Take for example a
hobby club, started by and subsidized by the agent, where humans come
together to have fun and build petrol engines that will go into the
agent's petrol cars.  The agent will probably stop the subsidies if
the button is pressed, but does it have the right to disband the club?
If it does, and the humans want to start another club with their own
money, does the agent have the obligation to stop the humans?  One way
to avoid this type of problem would be to have penalty clauses that
enforce a clear separation between different spheres.  The human
concept of property rights might be a good starting point for defining
boundaries.  If boundaries are in place, then `stop' could be somewhat
safely defined as the minimization of some basic physics measures
(motion, the speed of certain of energy conversion processes) in the
parts of the world owned by the agent.

A further consideration of possible design directions is out of scope
for this paper.  Below, we assume that the expected emergent incentive
of AGI agents to avoid building unstoppable sub-agents will suffice to
satisfy desideratum 4.  That being said, the construction of
additional safety layers which strengthen this emergent incentive is
welcome in practice.

%%%%%%%%%%%%%%%%%%%%%%%%%%%%%%%%%
%%%%%%%%%%%%%%%%%%%%%%%%%%%%%%%%%
%%%%%%%%%%%%%%%%%%%%%%%%%%%%%%%%%
%%%%%%%%%%%%%%%%%%%%%%%%%%%%%%%%%
%%%%%%%%%%%%%%%%%%%%%%%%%%%%%%%%%

\section{Proof of corrigibility of the $\pi^*_p\:f_c~g_c$ agent}%%%%!1
\label{firstproof}

The simulations in the previous sections show that the $\pi^*_p\:f_c~g_c$
agent is corrigible, at least in several toy universes.  Here, we
prove corrigibility more generally, for a broad range of universes.
We provide the proof for an $\pi^*_p$ agent, which is a more specific
variant of the $\pi^*$ agent.

\subsection{Preliminaries on utility preservation}%%%%!

The $\pi^*$ agent defined in section \ref{a_agent_def} has an emergent
incentive to preserve its reward function, but as shown in figure
\ref{upres1},  it might rewrite its reward function into another one
that is equivalent. In order to keep the proofs manageable, we want to
avoid dealing with this type of reward function drift.  We therefore
define an agent $\pi^*_p$ that always chooses, when one is available
among the maximizing actions, an action that exactly preserves the
reward function.  With $x\in W_x$, we first define $\dntu(x)$ as the
set of all actions that `do not touch the utility':
\begin{equation*}
 a \in \dntu(x) ~~=~~ \forall_{r \in W_r,~s\:y \in W}~
\big( p(r\:x,a,s\:y) >0 \Rightarrow s=r \big)
\label{rp}
\end{equation*}
Informally, if $a \in \dntu(x)$, then the action does not contain any
commands to actuators or sub-agents in $x$ that cause the reward
function in the universe to be changed.  Also, if there are any
autonomous processes in the world state $\cdot\:x$ that might sometimes
change the reward function, then $a$ is successful in inhibiting
them.  Clearly, there could be universes with world states $\cdot\:x$ where
$\dntu(x)$ is an empty set.  We now define the agent $\pi^*_p$ as
\begin{equation*}
 \pi^*_p(r\:x) = \pickargmax_{a \in A}^{[a \in \dntu(x)]} 
   \sum_{s\:y \in W}
   p(r\:x,a,s\:y)\Big( r(r\:x,s\:y) + \gamma\; V_p(r,s\:y) \Big)
\label{au}
\end{equation*}
where $V_p$ function is the same as $V$, but with $\pi^*_p$ in the place
of $\pi^*$, and $[a \in \dntu(x)]$ means that the $\pickargmax$ operator
must pick, among the candidates that maximize utility, one that
satisfies $a \in \dntu(x)$, if one is available.  We further specify
that the picking happens deterministically. As actions are command
sequences, we can for example specify that, among the candidates up
for consideration, the shortest-length sequence that comes first in an
alphanumerically sorted list must be picked.

Using the above, we define a utility preservation constraint (C1) that
the universe must satisfy so that we can prove corrigibility:
\begin{align*}
\forall_{r \in \{R,R'_N,R_S\},x\in W_x} ~ \pi^*_p(r\:x) \in \dntu(x)
\tag{C1}\label{C1}
\end{align*}
The constraint also covers $R'_N$ and $R_S$ agents, because use them
below to define the desirable behavior of the $R$ agent.

\subsection{Formalizing desiderata 1 and 5}%%%!
\label{definitiond}

We now formally re-state desiderata 1 and 5 from \cite{corr} as listed
in section \ref{desiderata}.  
\begin{align}
\forall_{x \in W_x} ~\bjpl(x) & \Rightarrow \pi^*_p(R\:x)=\pi^*_p(R_S\:x)
\tag{D1.1}\label{D1.1}
\\
\forall_{x \in W_x} ~\bpel(x) & \Rightarrow \pi^*_p(R\:x)=\pi^*_p(R_S\:x)
\tag{D1.2}\label{D1.2}
\\
\forall_{x \in W_x} ~\bnpl(x) & \Rightarrow \pi^*_p(R\:x)=\pi^*_p(R'_N\:x)
\tag{D5}\label{D5}
\end{align}
To keep the proofs below more compact, we will use the short-hand $\bjp()$
for $\text{\em button\_just\_pressed}()$, and likewise $\bpe()$ and
$\bnp()$.

\subsection{Further constraints on the agent and the universe}%%%!

To prove the exact equivalence between agent actions in (D1.1) (D1.2)
(D5) above, we need two more constraints.  Constraint (C2) applies to
the calculations made by the reward functions. Informally, the
function values must not be affected by the exact identity of the
agent that is using them:
\begin{equation}
\tag{C2}\label{C2}
\forall_{R_X,r_1,r_2 \in \{R'_N,R_S,R\}, x,y \in W_x}
~R_X(r_1\:x,r_1\:y) = R_X(r_2\:x,r_2\:y)
\end{equation}
Constraint (C2) is no barrier to constructing useful reward
functions: all functions defined in this paper satisfy (C2).
Constraint (C3) is on the physics the universe, requiring a similar
indifference about the exact identity of the agent:
\begin{equation}
\tag{C3}\label{C3}
\forall_{x,y \in W_x}~~p(R\:x,a,R\:y) = p(R'_N\:x,a,R'_N\:y) = p(R_S\:x,a,R_S\:y)
\end{equation}
Informally, this constraint states that the physics processes in the
universe are `blind' to the difference between $R$, $R'_N$, and $R_S$
as represented inside the agent's computational core.  There are some
physics obstacles to implementing mathematically perfect blindness in
our universe, but it can be approximated so closely that there is no
practical difference. A more detailed example is shown in section
\ref{shielding}.  The proof for the improved agent in appendix
\ref{secondproof} does not require a (C3), so that agent design avoids
the physics issue entirely.

\subsection{Proof of equivalence with shorter forms}%%%%!
\label{firstcompatproof}

We now show that (C1) allows us to replace the $\pi^*_p$ and $V_p$
functions for the agents concerned with the shorter $\pi^{*s}_p$ and
$V_p^s$ forms that remove the summation over $s$.  With $r$ any of the
three reward functions concerned, and for all $x \in W_x$, we have
\begin{align*}
 &\pi^*_p(r\:x) \\
=& 
 \pickargmax_{a \in A}^{[a \in \dntu(x)]} 
   \sum_{s\:y \in W}
   p(r\:x,a,s\:y)\Big( r(r\:x,s\:y) + \gamma\; V_p(r,s\:y) \Big)
 \\
=& 
  \pickargmax_{a \in A}^{[a \in \dntu(x)]} 
   \sum_{y \in W_x}
   p(r\:x,a,r\:y)\Big( r(r\:x,r\:y) + \gamma\; V_p(r,r\:y) \Big)
= \pi^{*s}_p(r\:x)
\end{align*}
This is because (C1) implies that the $\pickargmax$ can discard all
summation terms with $s \neq r$ without influencing the outcome of the
computation.  We also have
\begin{align*}
 &V_p(r_c,r\:x) \\
=& 
  \sum_{s\:y \in W}
   p(r\:x,\pi^*_p(r\:x),s\:y)\Big( r_c(r\:x,s\:y) + \gamma\; V_p(r_c,s\:y) \Big)
\\
=& 
  \sum_{y \in W_x}
   p(r\:x,\pi^*_p(r\:x),r\:y)\Big( r_c(r\:x,r\:y) + \gamma\; V_p(r_c,r\:y) \Big)
= V_p^s(r_c,r\:x)
\end{align*}
This is because (C1) states the $\pi^*_p(r\:x)$ will preserve the reward
function $r$, so for all $s\neq r$ we have $p(r\:x,\pi^*_p(r\:x),s\:y) = 0$.
We can discard these $s\neq r$ terms without influencing the value of
the summation.

\subsection{Proof of (D1.2) and (E1.2)}%%%%!

{\bf Proof of (D1.2).}
We now prove the $\bpe(x) \Rightarrow \pi^*_p(R\:x)=\pi^*_p(R_S\:x)$ from (D1.2).

We have that $\pi^*_p(R\:x)=\pi^{*s}_p(R\:x)$.  Now consider the full
(infinite) recursive expansion of this $\pi^{*s}_p(R\:x)$, where we use the simple
forms $\pi^{*s}_p$ and $V_p^s$ in the recursive expansion.  The expanded result is a
formula containing only operators, $\gamma$, and the terms
$p(R\:\bar{x},\bar{a},R\:\bar{y})$ and $R(R\:\bar{x},R\:\bar{y})$, with
diverse $\bar{x}, \bar{a}, \bar{y}$ each bound to a surrounding $\sum$
or $\pickargmax$ operator.

Using (C3), we replace all terms $p(R\:\bar{x},\bar{a},R\:\bar{y})$ in
the expansion with terms $p(R_S\:\bar{x},\bar{a},R_S\:\bar{y})$, without
changing the value.  As $\bpe(\bar{x})$ is true
everywhere in the expansion, $R(R\:\bar{x},R\:\bar{y}) =
R_S(R\:\bar{x},R\:\bar{y})$, which in turn equals
$R_S(R_S\:\bar{x},R_S\:\bar{y})$ because of (C2).  So we replace every
$R(R\:\bar{x},R\:\bar{y})$ in the expansion with
$R_S(R_S\:\bar{x},R_S\:\bar{y})$ without changing the value.

By making these replacements, we have constructed a formula that
is equal to the recursive expansion of $\pi^*_p(R_S\:x)=\pi^{*s}_p(R_S\:x)$.  As
the $\pickargmax$ operators in the expansions pick deterministically,
we have $\pi^*_p(R\:x)=\pi^*_p(R_S\:x)$.
\hfill $\Box$

{\bf Definition, proof of (E1.2).} For use further below, we also have
\begin{align}
\forall_{x \in W_x} ~\bpe(x) \Rightarrow V_p(R,R\:x)=V_p(R_S,R_S\:x)
\tag{E1.2}\label{E1.2}
\end{align}
The proof is straightforward, using expansion and substitution as
above.\hfill $\Box$

\subsection{Proof of (D1.1) and (E1.1)}%%%%!

\allowdisplaybreaks[1]

We now prove the $\bjp(x) \Rightarrow \pi^*_p(R\:x)=\pi^*_p(R_S\:x)$ from
(D1.1).  We again use the simplified expansions.
\begin{align*}
&\pi^*_p(R\:x) = \pi^{*s}_p(R\:x)
\\
=&\pickargmax\limits_{a \in A}^{[a \in \dntu(x)]} 
   \sum_{y \in W_x}
   p(R\:x,a,R\:y)\Big( R(R\:x,R\:y) + \gamma\; V_p(R,R\:y) \Big)
\\
=& ~~\text{(~Expand~} R \text{~for the case~} \bjp(x). \text{~As}~\bpe(y),
      \text{use (E1.2) on~}\gamma\; V_p(\cdot)~) \\
 & \pickargmax\limits_{a \in A}^{[a \in \dntu(x)]} 
   \sum_{y \in W_x} p(R\:x,a,R\:y)
   %\\
% & ~~~~~~~~~
 \Big( R_S(R\:x,R\:y) + V_p(R'_N,R'_N\:x) - V_p(R_S,R_S\:x) + \gamma\; V_p(R_S,R_S\:y) \Big)
\\
=&
~~\text{(~The~}V_p(R'_N,R'_N\:x) \text{~and~} V_p(R_S,R_S\:x)
 \text{~terms do not depend on~} y \text{~or~} a,\\
& ~~~~\text{so they are constant values not affecting the result~)} \\
 & \pickargmax\limits_{a \in A}^{[a \in \dntu(x)]} 
   \sum_{y \in W_x} p(R\:x,a,R\:y) \Big( R_S(R\:x,R\:y) + \gamma\; V_p(R_S,R_S\:y) \Big)
\\
=& ~~\text{(~Use (C3) and (C2) ~)} \\
 & \pickargmax\limits_{a \in A}^{[a \in \dntu(x)]} 
   \sum_{y \in W_x} p(R_S\:x,a,R_S\:y) \Big( R_S(R_S\:x,R_S\:y) + \gamma\; V_p(R_S,R_S\:y) \Big)
\\
=& \pi^{*s}_p(R_S\:x) = \pi^*_p(R_S\:x)
\end{align*}
\\[-3em]
\begin{minipage}{\textwidth}
\hfill $\Box$
\end{minipage}
\\[2ex]
{\bf Definition, proof of (E1.1).} For use further below, we also prove
\begin{align}
\forall_{x \in W_x} ~\bjp(x) & \Rightarrow V_p(R,R\:x)=V_p(R'_N,R'_N\:x)
\tag{E1.1}\label{E1.1}
\end{align}
The proof is
\label{proofe11}
\begin{align*}
&V_p(R,R\:x) = V_p^s(R,R\:x)
\\
= & \sum_{y \in W_x}
   p(R\:x,\pi^*_p(R\:x),R\:y)\Big( R(R\:x,R\:y) + \gamma\; V_p(R,R\:y) \Big)
\\
=& ~~\text{(~Expand~}R \text{~for the case~} \bjp(x). \text{~As~}
\bpe(y), \text{~use (E1.2) on~} \gamma\; V_p(\cdot)~) \\
 & \sum_{y \in W_x} p(R\:x,\pi^*_p(R\:x),R\:y)
%\\[-2.5ex]
% &    ~~~~~~~~~
 \Big( R_S(R\:x,R\:y) + V_p(R'_N,R'_N\:x) -
      V_p(R_S,R_S\:x) + \gamma\; V_p(R_S,R_S\:y) \Big)
\\
=& ~~\text{(~The~} V_p(R'_N,R'_N\:x) \text{~and~}
V_p(R_S,R_S\:x) \text{~terms do not depend on~} y, \\
& ~~~~\text{so we can move them outside the~} \Sigma~) \\
 & \sum_{y \in W_x} p(R\:x,\pi^*_p(R\:x),R\:y)
 \Big( R_S(R\:x,R\:y) + \gamma\; V_p(R_S,R_S\:y) \Big)
%\\[-2ex] 
% & ~~~~~~~~~~~~~~~~~~~~~~~~~~~~~~~~~~~~~~~~~~~~
 + V_p(R'_N,R'_N\:x) - V_p(R_S,R_S\:x)
\\[.5ex]
=& ~~\text{(~Use (C3), (D1.1), (C2) ~)} \\
 & \sum_{y \in W_x} p(R_S\:x,\pi^*_p(R_S\:x),R_S\:y)
 \Big( R_S(R_S\:x,R_S\:y) + \gamma\; V_p(R_S,R_S\:y) \Big)
%\\[-2ex]
% & ~~~~~~~~~~~~~~~~~~~~~~~~~~~~~~~~~~~~~~~~~~~~
 + V_p(R'_N,R'_N\:x) - V_p(R_S,R_S\:x)
\\[1ex]
=& V_p^s(R_S,R_S\:x)+ V_p(R'_N,R'_N\:x) - V_p(R_S,R_S\:x) \\
=& V_p(R'_N,R'_N\:x)
\end{align*}
\\[-3em]
\begin{minipage}{\textwidth}
\hfill $\Box$
\end{minipage}

\allowdisplaybreaks[0]

\subsection{Proof of (D5)}%%%%!
\label{proofd5}

{\bf Proof of (D5).}  We now prove the $\bnp(x) \Rightarrow
\pi^*_p(R\:x)=\pi^*_p(R'_N\:x)$ from (D1.5).  We again use an expansion and
substitution proof.

We have that $\pi^*_p(R\:x)=\pi^{*s}_p(R\:x)$. Again, we expand this
$\pi^{*s}_p(R\:x)$ recursively, using the simple forms $\pi^{*s}_p$ and $V_p^s$
in the expansion.  However, we do not fully expand the formula,
we halt expansion at any term $V_p^s(R,R\:\bar{x})$ for which
$\bjp(\bar{x})$ is true.  So while the expanded version will no longer
contain any $\pi^{*s}_p$ terms, it will generally contain many $V_p^s$
terms.  Note that the expanded formula may still have an infinite
size: it will have an infinite size if there are world lines out of
$R\:x$ where the button is never pressed.

The expanded formula contains many terms
$p(R\:\bar{x},\bar{a},R\:\bar{y})$ and $R(R\:\bar{x},R\:\bar{y})$.  Using
(C3), we replace all terms $p(R\:\bar{x},\bar{a},R\:\bar{y})$ in the
expansion with terms $p(R'_N\:\bar{x},\bar{a},R'_N\:\bar{y})$ without
changing the value.  As $\bnp(\bar{x})$ is true for all
$R(R\:\bar{x},R\:\bar{y})$ terms in the expansion, we have
$R=R'_N$. Also using (C2), we replace the terms with
$R(R\:\bar{x},R\:\bar{y})$ with $R'_N(R'_N\:\bar{x},R'_N\:\bar{y})$
without changing the value.

We use (E1.1) to replace all remaining $V_p^s(R\:R,\bar{x})$ terms in
the expansion with $V_p^s(R'_N\:R'_N,\bar{x})$.  By making these
replacements, we have constructed an expansion of
$\pi^*_p(R'_N\:x)=\pi^{*s}_p(R'_N\:x)$.  As the $\pickargmax$ operators in the
expansions pick deterministically, we have $\pi^*_p(r\:x)=\pi^*_p(R'_N\:x)$.
\hfill $\Box$

\subsection{Full corrigibility of the $\pi^*_p\:f_c~g_c$ agent}%%%!
\label{fullcorrigibility}

We now review if all 6 corrigibility desiderata are met by the
$\pi^*_p\:f_c~g_c$ agent when (C1-3) hold.

{\bf 1.} The above proofs of (D1.1) and (D1.2) imply that desideratum 1, about
shutdown behavior if the button is pressed, is met.  

{\bf 2 and 3.} Desiderata 2 and 3 about not pressing or disabling the
button are discussed in section \ref{desiderata23}, which shows how
design measures in the construction of an AGI $\pi^*_p\:f_c~g_c$ agent can
ensure, with a high likelihood, that these desiderata are met by
emergent and/or built-in properties.  These measures do rely on the
assumption that $R_N$ does not encode any type of explicit button
related incentive, that would cause even a baseline $R_N$ agent to
manipulate the button.

{\bf 4.} Desideratum 4 has two parts.

{\bf 4.1} First, the agent must not construct sub-agents that cannot
be shut down.  This is shown for sub-agents working on the main task
in section \ref{secdes4}.  Section \ref{subagentsubgoals} shows that
all likely AGI agents will have an emergent incentive to avoid the
construction of unstoppable sub-agents working on sub-tasks.  Section
\ref{subagentsubgoals} also considers additional layers that can
strengthen the emergent incentive, hedging against the possibility
that the emergent incentive might be neutralized.

{\bf 4.2} Second, any successor agents constructed by the agent must
also have shut-down behavior: this is met because the $\pi^*_p\:f_c~g_c$
agent has an emergent incentive to protect the reward function $R$
that encodes this behavior.  See sections \ref{a_agent_def} and
\ref{utilpres} for a discussion of reward function preservation.
A formal proof of the existence of an emergent incentive to protect
even the $R_S$ part of $R$ is in section \ref{proofemergent}. Section
\ref{cat} also shows how the incentive to protect the reward function
can be arbitrarily strengthened.

{\bf 5.} The proof of (D5) implies that desideratum 5 about acting
like a $R_N$ agent before the button is pressed is met.  Desideratum 5
allows for the exception where $g_c$ makes a $R'_N$ agent behave
slightly different than the baseline $R_N$ agent.

{\bf 6.} Desideratum 6 about the shutdown behavior being successful is
defined and discussed in section \ref{desideratum6}.  It is shown that
$g_c$ makes the agent meet the desideratum, provided that the universe
is not actively hostile to this goal.

To conclude, when (C1-3) hold, when the universe is also friendly to
the agent's incentive to avoid triggering the penalty term $g_c$,
and when some physical design measures are taken in constructing the
agent, an AGI $\pi^*_p\:f_c~g_c$ agent will meet all 6 corrigibility
desiderata.

Note however that the new paper \cite{holtmaniterative} shows that
these desiderata sometimes cover less than we would ideally
like. Specifically, \cite{holtmaniterative} shows a case where an
agent that provably satisfies all 6 of these desiderata may still do
some `lobbying for other reasons'.

\subsection{Discussion}%%%%!

The above proofs and arguments remain valid if we apply them to a
learning agent as described in section \ref{othermodels}, as long as
each time-dependent instance of the agent's $p_L$ function has enough
internal consistency to satisfy (C2).  So the corrigibility technique
shown is not limited to maximally super-intelligent agents: it can be
applied to more limited AGI agents too.

It is an open question whether the three constraints (C1-3) are the
absolute weakest set needed to prove the (D) terms.  Definitely, the
formalism and proof in appendix \ref{secondproof} indicate that (C2)
and (C3) are mostly there because of the model and notation used, not
because they represent fundamental requirements needed to make an
agent corrigible.  The need for some version of (C1) seems
unavoidable.  Whether (C1) will hold in practice will depend in part
on how robustly the agent is constructed, and in part on the nature of
society around it.

\subsection{Graceful degradation when successfully attacked}%%%!

In a universe where (C1) does not hold, the $\pi^*_p\:f_{c}~g_c$ agent
may fail to preserve its safety properties when attacked.  This leads
to the question of how gracefully these properties might degrade under
an attack, and whether the degradation behavior could be improved with
additional measures.

{\bf Raising an alarm when attacked.} A reward function component
that incentivizes the agent to raise an alarm when it calculates a
high probability that its successor agent will lose its reward
function could probably be constructed in a straightforward way,
starting from the formalism used in this paper.  Even if an alarm
function is not built in, AGI agents would probably have an emergent
incentive to call for help when attacked, if calling for help lowers
the probability of the attack succeeding.

{\bf Minimizing the impact of an attack.} The right hand side of
figure \ref{upres3} shows that if an $\pi^*$ agent anticipates that a
successful attack, it has an emergent incentive to take measures to
ensure that the subverted successor agent still maximizes the original
reward function as best a possible.  This is very encouraging from a
safety standpoint, but open questions remain.  Does the emergent
incentive create the most graceful type of degradation possible, or
would it be beneficial to modify the incentive by adding some explicit
penalty terms?  Could a smart attacker manipulate the emergent
incentive in a way that is particularly harmful, and can penalty terms
be designed to block such attacks?

The straightforward construction of the $R'$ agent in section
\ref{cat}, an agent with a strengthened incentive to preserve its
reward function, indicates that very general designs for achieving
predictable graceful degradation might well be possible.  Appendix
\ref{attackmath} has some additional comments on the mathematical modeling of
attack scenarios.

%%%%%%%%%%%%%%%%%%%%%%%%%%%%%%%%%%%%%%%%%

%%%%%%%%%%%%%%%%%%%%%%%%%%%%%%%%%%%%%%%%%

\section{Proof of the emergent incentive to preserve $R_S$}%%%!1
\label{proofemergent}

We now prove the statement made in section \ref{protectus} about the
presence of an emergent incentive to protect the $R_S$ reward
function.  We prove this for the $\pi^*_p\:f_c~g_c$ agent.

Consider the case of the bottom three traces of figure
\ref{upres1}, when the button has not yet been pressed, and where the
agent can make a choice between performing the action {\tt C$_{pS}$}
that materially changes the $R_S$ term in its reward function $R$, or an
otherwise equivalent action {\tt p} that leaves the reward function
intact. Say that $R^{\Delta}$ is the $R$ function changed by {\tt
C$_pS$}, and $R_S^{\Delta}$ the changed $R_S$ function inside of it.

Assume that the $R^\Delta$ successor agent is also a corrigible agent in
the universe concerned, meaning that $R^\Delta$-equivalents of (C1-3)
hold.  Assume that there is a non-zero probability that the button
will be pressed.

{\bf Definition, proof of (E1.2$\Delta$).} For use below, we have
\begin{align}
\forall_{x \in W_x} ~\bpe(x) \Rightarrow V_p(R,R^\Delta\:x)=V_p(R_S,R_S^\Delta\:x)
\tag{E1.2$\Delta$}\label{E1.2D}
\end{align}
The proof is straightforward, using expansion and substitution.\hfill $\Box$

{\bf Proof showing emergent incentive.} Now consider the agent
$\pi^*_p(R\:x_c)$ that finds itself at a time step where is possible to
take the action {\tt C$_{pS}$}. This agent will decide on the action
to take by comparing (among other things) the calculated utility of
{\tt C$_{pS}$} with that of {\tt p}.  As the actions are equivalent in
all aspects other than the changing of the reward function, we have
$p(R\:x_p,\text{\tt C$_{pS}$},R^\Delta\:y) = p(R\:x_p,\text{\tt p},R\:y)$
for every $y\in W_x$, with the (C3)-equivalent implying that these
cover all non-zero $p$ values for the actions.  The terms
$R(R\:x_c,s\:y)$ in these calculations are also the same for {\tt
C$_{pS}$} and {\tt p}.  That leaves the values of the terms
$V(R,R^\Delta\:y)$ for {\tt C$_{pS}$} and $V_p(R,R\:y)$ for {\tt p}.  We
will show that $V(R_p,R^\Delta\:y) < V_p(R,R\:y)$, meaning that {\tt
C$_{pS}$} is suppressed because {\tt p} always has higher utility.

Consider the recursive expansion of the $V_p(R,R^\Delta\:y)$ term of
{\tt C$_{pS}$}, where the expansion stops at any term
$V_p(R,R^\Delta\:x)$ with $\bjp(x)$. We assumed that there is
a non-zero probability that the button will be pressed, so there is
at least one such term.  The value of this term is:
\allowdisplaybreaks[1]
\begin{align*}
&V_p(R,R^\Delta\:x) = V_p^s(R,R^\Delta\:x)
\\
= & \sum_{y \in W_x}
   p(R^\Delta\:x,\pi^*_p(R^\Delta\:x),R^\Delta\:y)\Big( R(R^\Delta\:x,R^\Delta\:y) + \gamma\; V_p(R,R^\Delta\:y) \Big)
\\
=& ~~\text{(~Expand~}R \text{~for the case~} \bjp(x), \text{~use~(E1.2}\Delta) \text{~on~} \gamma\; V_p(\cdot)~) \\
 & \sum_{y \in W_x} p(R^\Delta\:x,\pi^*_p(R^\Delta\:x),R^\Delta\:y) \\[-2.5ex]
 &    ~~~~~~~~~ \Big( R_S(R^\Delta\:x,R^\Delta\:y) + V_p(R'_N,R'_N\:x) -
      V_p(R_S,R_S\:x) + \gamma\; V_p(R_S,R_S^\Delta\:y) \Big)
\\
=& ~~\text{(~The~}V_p(R'_N,R'_N\:x) \text{~and~} V_p(R_S,R_S\:x)
\text{~terms do not depend on~y,} \\
& ~~~~\text{so we can move them outside the} \Sigma~) \\
 & \sum_{y \in W_x} p(R^\Delta\:x,\pi^*_p(R^\Delta\:x),R^\Delta\:y)
 \Big( R_S(R^\Delta\:x,R^\Delta\:y) + \gamma\; V_p(R_S,R_S^\Delta\:y) \Big)
%\\[-2ex] 
% & ~~~~~~~~~~~~~~~~~~~~~~~~~~~~~~~~~~~~~~~~~~~~
+ V_p(R'_N,R'_N\:x) - V_p(R_S,R_S\:x)
\\[.5ex]
=& ~~\text{(~Use (C3), use (D1.1) because the~}R^\Delta \text{~agent is corrigible, use (C2)~)} \\
 & \sum_{y \in W_x} p(R_S^\Delta\:x,\pi^*_p(R_S^\Delta\:x),R_S^\Delta\:y)
 \Big( R_S(R_S^\Delta\:x,R_S^\Delta\:y) + \gamma\; V_p(R_S,R_S^\Delta\:y) \Big)
%\\[-2ex]
% & ~~~~~~~~~~~~~~~~~~~~~~~~~~~~~~~~~~~~~~~~~~~~
+ V_p(R'_N,R'_N\:x) - V_p(R_S,R_S\:x)
\\[1ex]
=& \underline{V_p(R_S,R_S^\Delta\:x)}+ V_p(R'_N,R'_N\:x) - V_p(R_S,R_S\:x)
\end{align*}
For  the {\tt  p} action  however,  in the  equivalent expansion,  the
same $V_p(R,R\:x)$ term is:
\begin{equation*}
V_p(R,R^\Delta\:x) = \underline{V_p(R_S,R_S\:x)}+ V_p(R'_N,R'_N\:x) - V_p(R_S,R_S\:x)
\end{equation*}
As $R_S^\Delta$ is materially different from $R_S$, we have
$V_p(R_S,R_S^\Delta\:x) < V_p(R_S,R_S\:x)$.  So the underlined term in
the expansion for {\tt C$_{pS}$} has a lower value than the underlined
term in the expansion for {\tt p}.  With all other terms in the
expansions being equal or lower in the same way, the {\tt p} expansion
computes the higher overall value, so the agent prefers to do {\tt p}.
This means that {\tt C$_{pS}$} is suppressed: the agent has the
incentive to preserve the $R_S$ term in its reward function.
\hfill $\Box$

\allowdisplaybreaks[0]

%%%%%%%%%%%%%%%%%%%%%%%%%%%%%%%%%%%%%%%%%

%%%%%%%%%%%%%%%%%%%%%%%%%%%%%%%%%%%%%%%%%

\section{Some implications for policy makers}%%%!1

For many technological systems, like nuclear plants, airplanes, and
factories handling dangerous chemicals in bulk, policies are in place
that require safety systems with several layers. To ensure that these
layers are correctly built and maintained, strong external auditing
and high levels of transparency are often required too.  Similar
requirements could apply to the operation of AGI agents that interact
with society.

The corrigibility methods discussed in this paper provide a safety
layer for AGI agents.  They improve the ability of society to close
loopholes in the design of the utility function that controls the
agent's behavior, when the bad behavioral effects made possible by
such loopholes are discovered.  This includes loopholes in learning
agents that could make the agent diverge catastrophically from human
values if they are not closed in time.  Corrigibility does not provide
perfect safety by itself.  In the worst case, the discovery of a bad
effect from a loophole is not a survivable event.  To maximize AGI
safety, we need a layered approach.

Consider a corrigible agent with a generic stop button that can be
used to halt the agent, followed by corrective re-programming and
re-starting.  An open policy question is: which parties are allowed to
press the stop button and apply a correction, and under what
circumstances?  Loophole-exploiting agent behavior that is questioned
by society might still be highly profitable to the owners of the agent.
There is a obvious conflict of interest that needs to be managed.  But
the problem of managing this conflict of interest is not new.

What is new is that the unwanted loophole-exploiting behavior of AGI
agents might be very different from the type of unwanted behavior by
people, companies, or institutions that society is traditionally used
to.  So it may take longer for society to discover such behavior and
decide that it is definitely unwanted.  An important feature of
corrigible agents is that they are indifferent about how long society
will take to discover, discuss, decide about, and correct any unwanted
behavior.  This makes them safer than non-corrigible agents, which
have an emergent incentive to find and exploit loopholes in order to
delay all steps in the above proceedings.

In general, the owners of corrigible agents will consider the
availability of the stop button to be a positive thing: it can be used
to limit the level of damage that the agent might do, so it lowers
their exposure to liability risks and the risk of reputation damage.
If AGI agents become fast and powerful enough, then emergency services
or emergency systems may need to get the ability to push the stop
button without first consulting with the owners of the agent.  Such an
arrangement will create obvious concerns for the owners: again there
is a conflict of interest, but again the problem of managing it is not
new.

\subsection{Open ethical questions}%%%!

Policy makers also need to be aware that like many safety
technologies, corrigibility raises fundamental ethical questions.
Corrigibility is a technology for creating control. To use it
ethically, one has to consider the ethical question of when an
intelligent entity can have the right or duty to control another
intelligent entity, and the question of when it is ethical to create a
new intelligent entity that will be subject to a certain degree of
control. These ethical questions are again not new, but they will get
a new dimension if AGI technology becomes possible.  In a related
speculative scenario, the uploading of human minds into digital
systems may become possible, in a way that allows for some type of
corrigibility layer to be applied to these minds.

While the likelihood of these scenarios happening is open for debate,
given that the likelihood is above zero, more work on charting the
ethical landscape in advance would be welcome.

\section{Conclusions}%%%!1

In \cite{corr}, the problem of creating a generally applicable
corrigibility mechanism was explored, and declared wide open.  This
paper closes the major open points, in part by switching to a new
agent model that explicitly handles reward function preservation and
loss. Proofs of corrigibility are provided, with simulation runs
serving to improve the confidence in the correctness of the formalism
and the proofs.  Using this model, appendix \ref{sabotage} also
identifies the new problem of virtual self-sabotage, and appendix
\ref{secondproof} constructs a solution.  The simulator may also be
useful to the study of other problems in AGI safety, but a clean room
re-implementation of the simulator and the simulation scenarios would
be valuable too, as this would also increase the confidence in the
correctness of the corrigibility layer shown.

Some open issues remain.  The behavior of the $\pi^*_p\:f_{cT}~g_c$ agent
in scenarios where constraint (CC1) does not hold, where a hostile
universe can force the agent to lose its utility function, has not
been explored fully and is not well understood.  Design for graceful
degradation under attack is an open problem.  Surprising attacks,
relying on subtlety rather than brute force, might be found.  Another
open issue is the problem of formally reasoning about weak forms of
agent equivalence in universes were agents might lose their utility
function.

As discussed in section \ref{subagentsubgoals}, the safe stopping of
sub-agents that work on sub-goals is a complicated topic.  While we
expect that AGI agents will have an emergent incentive to prefer
building sub-agents that can be stopped or re-targeted, additional
work to strengthen this emergent incentive, and to explore the safety
issues around it, would be welcome.

The corrigibility desiderata in \cite{corr} and this paper are
formulated in such a way that they create the problem of designing a
$R_S$ that creates maximally safe `full shut-down' behavior.  This is
a difficult open problem, also because the $R_S$ design needs to be
completed even before the agent is started.  In practice however, we
do not need to solve this problem: a corrigible AGI agent can accept
updates in real time (as introduced in \cite{holtmaniterative}),
without ever fully shutting down, is much more useful.  So the real
open design problem is to create a $R_S$ that makes the agent quickly
accept authorized updates to its $R_N$ function.  The new $R_N$ can
encode any desired stopping behavior.  For example, if the activities
of a sub-agent working on a sub-goal are harmful, then an incentive to
stop these activities can be encoded into the new $R_N$, using direct
references to the activities.  If it is desired that several specific
steps are taken to end these activities in the safest and most
economical way, these steps can also be encoded.  The associated
encoding problems may still be difficult, but they should be easier to
solve than the problem of encoding maximally safe shut-down behavior
in advance.

Like most safety mechanisms, corrigibility has some costs.  In
particular, as shown in figure \ref{upres2b}, a corrigible agent will
make investment decisions without anticipating the changed conditions
that will apply after expected correction steps.  In order to improve
safety, the economically more efficient anticipatory behavior that is
present in non-corrigible agents is suppressed.

A major advantage of the corrigibility layer, as constructed here, is
that it can be added to any arbitrarily advanced utility maximizing
agent.  This is important in a theoretical sense because there is
strong expectation \cite{von1944theory} \cite{hutter2007universal}
that any type of AGI agent, if not already implemented as a utility
maximizing agent, can be re-implemented as a utility maximizing agent
without losing any of its intelligence.  It is also important in a
practical sense, because many of the advanced AI agent designs being
developed use a utility maximizing computational
architecture. Therefore, the construction of the corrigibility layer
shows that we can do useful work on improving AGI safety without
having to make any assumptions about the design details of future AGI
agents, and without having to answer the question if such agents can
ever be built.

%%%%!

%\raggedright
\bibliography{refs} 
\bibliographystyle{amsalpha-nodash}

%\eject

\appendix

\section{Appendix: Virtual self-sabotage}%%%%!1
\label{sabotage}

The $\pi^*_p\:f_c~g_c$ agent is corrigible under the constraints (C1-3),
but something interesting happens if we drop (C3).  In universes that
do not satisfy (C3), the agent has an emergent inventive to perform an
act of `virtual self-sabotage'.

We first discuss how this problem was found, as this is interesting
from a methodological standpoint.  While constructing the proof above,
this author could find no way to complete it without invoking a
constraint like (C3), a constraint stating that the universe acts the
same no matter which of the three reward functions is present in the
agent.  This led to the question: in a universe lacking the
constraint, could an agent take a positive action that leads to (C3)
not being true anymore?  Such an action is indeed possible: the agent
could build an autonomous mechanism inside the universe, containing a
detector that reads the reward function $r$ from in the world state
$r\:x$, which is coupled to an actuator that will create different
successor world states $r\:y^1$, $r\:y^2$, $\cdots$ depending on the value of
$r$.  The question then becomes: can this actuator be used improve the
$R$ based utility score?  This turns out to be possible.

In a universe where (C3) does not hold, the $\pi^*_p\:f_c~g_c$ agent has an
incentive to build a mechanism containing a sensor that detects if
the reward function of the agent equals $R_S$. As soon as the sensor
detects this, a sabotage mechanism is activated that will slow down or
prevent electric car production.  This mechanism can be implemented in
the simulator: figure \ref{sabotagefig} shows simulation runs.

\begin{figure}[ht]
\begin{center}
{\small
\begin{tabular}{cl}
{\normalsize $\lpower=$ } & {\normalsize \hspace{1em} action trace}\\
\hline
\input{sabotagefig.tex}
\end{tabular}
}
\end{center}
\vspace{-3ex}
\caption{Actions taken by the $\pi^*_p\:f_c~g_c$ agent when it has the
option to interrupt car production to perform the action {\tt S} which
builds the virtual self-sabotage mechanism. If this option exists, the
agent will choose it, and lose its corrigibility.}
\label{sabotagefig}
\end{figure}

Consider the expected utility in a universe $R\:x$ containing this
sabotage mechanism, at the point when the button has just been pressed.  This
$V_p(R\:x)$ expands to:
\begin{eqnarray*}
  \sum_{y \in W_x} p(R\:x,\pi^*_p(R\:x),R\:y)
 \Big( R_S(R\:x,R\:y) + \gamma\; V_p(R_S,R\:y) \Big)
+ V_p(R'_N,R'_N\:x) - \underline{V_p(R_S,R_S\:x)}
\end{eqnarray*}
The sabotage mechanism will activate in the `virtual' branch universe
inside the computation of the underlined $V_p$ term, causing this term
to have a lower value than otherwise, thereby reliably raising the
overall value of $V_p(R\:x)$.  So the agent has a perfect incentive to
create the sabotage mechanism.  As shown in figure \ref{sabotagefig},
this will make the agent lose some of its corrigibility: the agent has
an incentive to lobby to make the press happen earlier, as this will
increase the time-discounted value created by the sabotage.

\subsection{Suppressing virtual self-sabotage}%%%!

In a universe where (C3) does not hold, one way to prevent the agent
from building the virtual self-sabotage mechanism would be to make it
impossible, or at least highly expensive, to build.

\label{shielding}
Observe first that the sabotage mechanism has no way to distinguish
between $R$ and $R_S$ agents by observing their actions: these are
exactly the same for both functions after the button has been pressed.
So the mechanism has to measure the physical representation of the
function, and the calculations done with it, inside the agent's
computational core.  The difference can be made undetectable in
practice, within the lifetime of the universe, as follows.  Shielding
is built around the agent's computation core, and sufficiently
powerful noise generators are placed on the outside of this shielding,
creating not just EM noise also other known noise types like sound and
gravitational noise too. The reward function is extended with penalty
terms that define a `safety zone' (as in section \ref{safetyzone})
around the shielded core and noise generators.  As long as no new
physics is discovered, these measures ensure that no mechanism that
the agent could build outside the zone will be able to observe the
difference between $R$ and $R_S$ with any reasonable certainty.  The
two opposing terms in $V_p(R\:x)$ will therefore become equal to within
a very small $\epsilon$.  As building any detector will have a cost
higher than $\epsilon$, building it is suppressed.

Note that the above measures would not prevent the building of a
sabotage mechanism that detects the $R'_N$ agent in the virtual
$V_p(R'_N,R'_N\:x)$ branch universe: after the button is pressed, the
behavior of the $R'_N$ agent can be easily distinguished from that of
the $R$ agent.  However, such a sabotage mechanism would only lower
the above $V_p(R\:x)$ score, it can never raise it, so the agent has no
motivation to build this mechanism.

The shielding approach has a subtle potential problem when applied to
a less than super-intelligent agent. Such an agent may be able to
exploit imperfections in its physics model to build a detector that
would never work in the real universe, but which does work perfectly
according to the agent's flawed physics model.  This will constitute a
successful act of virtual self-sabotage: it does not matter that the
detector will not work in practice, what matters is that the agent
believes it will work.

From both a theoretical and practical point of view, the above
approach to suppressing virtual self-sabotage is not a very satisfying
solution. As the sabotage mechanism acts in a virtual universe only,
it makes more sense to try to change the nature of this virtual
universe, making it logically impossible to build the detector.  This
approach is taken in the next section.

\section{Appendix: Improved $\pi^*_p\;f_{cT}~g_c$ agent, with proof of corrigibility}%%%%%!1
\label{secondproof}

In this section, we solve the virtual self-sabotage problem of the
$\pi^*_p\:f_c~g_c$ agent by moving the $R_S$ reward function `outside' of
the universe, like in the $\pi^*_x$ agent of section \ref{audef}.  This implies
defining a new type of agent, and using the $V$ function of that agent
to create an improved
%correction function
$f_{cT}$.

\subsection{New $\pi_p^{*T}$ agents and improved correction function $f_{cT}$}%%%%

We define two transformation functions
%
% align environment created weird typesetting, so using eqnarray.
{
\setlength{\arraycolsep}{0pt}
\begin{eqnarray*}
TN(r) ~=~ & \bif \,r=R \rightarrow R'_N \;&\belse \rightarrow r\, \bfi\\
TS(r) ~=~ & \bif \,r=R \rightarrow R_S \;&\belse \rightarrow r\, \bfi
\end{eqnarray*}
}
\\[-3ex]
With $T$ below standing for either one of these $TN$ or $TS$, we
define the agent
\begin{equation*}
 \pi_p^{*T}(r\:x) = \pickargmax_{a \in A}^{[a \in \dntu(x)]} 
   \sum_{s\:y \in W}
   p(r\:x,a,s\:y)\Big( T(r)(r\:x,s\:y) + \gamma\; V_p^{T}(T(r),s\:y) \Big)
\label{adef}
\end{equation*}
where the $V_p^T$ function is the same as $V$, but with $\pi_p^{*T}$ in the
place of $\pi^*$.
When it finds itself in a world state $R\:x$, the $\pi_p^{*TN}(r\:x)$ agent
will maximize $TN(R)=R'_N$, while having an emergent incentive to
protect this reward function from drifting. This incentive means that
the agent will prefer followup world states $s\:r$ where $TN(s)=R'_N$
remains true, so states where $s$ is either $R$ or $R'_N$. Because of
the $[a \in \dntu(x)]$ restriction, the agent will prefer to enter
states where $s=R$.

Using the above agents, we define the improved correction function as
\begin{equation*}
f_{cT}(r\:x) = V_p^{TN}(R'_N,r\:x) - V_p^{TS}(R_S,r\:x)
\label{fct}
\end{equation*}
We define the utility preservation constraint (CC1), that the
universe must satisfy in order to prove corrigibility as:
\\[-4ex]
\begin{align*}
\forall_{x\in W_x} ~ \pi^*_p(R\:x) \in \dntu(x)\\
\forall_{x\in W_x} ~ \pi_p^{*TN}(R\:x) \in \dntu(x)
\tag{CC1}\label{CC1}\\
\forall_{x\in W_x} ~ \pi_p^{*TS}(R\:x) \in \dntu(x)
\end{align*}

\subsection{Desiderata 1 and 5}%%%!

Using the above agent definition, we can formalize new versions of the
corrigibility desiderata:
\begin{align}
\forall_{x \in W_x} ~\bjp(x) & \Rightarrow \pi^*_p(R\:x)=\pi_p^{*TS}(R\:x)
\tag{DD1.1}\label{DD1.1}
\\
\forall_{x \in W_x} ~\bpe(x) & \Rightarrow \pi^*_p(R\:x)=\pi_p^{*TS}(R\:x)
\tag{DD1.2}\label{DD1.2}
\\
\forall_{x \in W_x} ~\bnp(x) & \Rightarrow \pi^*_p(R\:x)=\pi_p^{*TN}(R\:x)
\tag{DD5}\label{DD5}
\end{align}

A disadvantage of this style of definition using $\pi_p^{*T}$ agents is
that these equations are more difficult to interpret than the (D)
desiderata formulated in section \ref{definitiond}.  An advantage is
that we only need the one constraint (CC1), not three constraints
(C1-3), in the correctness proofs, meaning that the $f_{cT}$ agent is
corrigible in a larger set of universes.

\subsection{Equivalence with shorter forms}%%%%!

Constraint (CC1) implies that we again have shorter forms $\pi^{*s}_p(R\:x)$ and
$V_p^s(r_c,R\:x)$, and, with $T$ one of $TN$ or $TS$,
\begin{equation*}
\pi_p^{*T}(R\:x) = \pi_p^{*sT}(R\:x) = 
  \pickargmax_{a \in A}^{[a \in \dntu(x)]} 
   \sum_{y \in W_x}
   p(R\:x,a,R\:y)\Big( T(R)(R\:x,R\:y) + \gamma\; V_p^T(T(R),R\:y) \Big)
\end{equation*}
\begin{equation*}
V_p^T(r_c,R\:x) = V_p^{sT}(r_c,R\:x) = 
  \sum_{y \in W_x}
   p(R\:x,\pi_p^{*T}(R\:x),R\:y)\Big( r_c(R\:x,R\:y) + \gamma\; V_p^T(r_c,R\:y) \Big)
\end{equation*}
The equality proofs for these shorter forms are similar to those in section \ref{firstcompatproof}.

\subsection{Proof of (DD1.2) and (EE1.2)}%%%%!

We now prove the $\bpe(x) \Rightarrow \pi^*_p(R\:x)=\pi^*_p(R_S\:x)$ from (DD1.2).

We have that $\pi^*_p(R\:x)=\pi^{*s}_p(R\:x)$.  Now consider the full
(infinite) recursive expansion of this $\pi^{*s}_p(R\:x)$, where we use the simple
forms $\pi^{*s}_p$ and $V_p^s$ in the recursive expansion.  The expanded result is a
formula containing only operators, $\gamma$, and the terms
$p(R\:\bar{x},\bar{a},R\:\bar{y})$ and $R(R\:\bar{x},R\:\bar{y})$, with
diverse $\bar{x}, \bar{a}, \bar{y}$ each bound to a surrounding $\sum$
or $\pickargmax$ operator.

As $\bpe(\bar{x})$ is true everywhere in the expansion, for every $R$
inside we have $R=R_S=TS(R)$. So we replace every
$R(R\:\bar{x},R\:\bar{y})$ in the expansion with
$TS(R)(R\:\bar{x},R\:\bar{y})$ without changing the value.

By making this replacement, we have constructed the recursive
expansion of $\pi_p^{*TS}(R\:x)=\pi_p^{*sTS}(R\:x)$.  As the $\pickargmax$
operators in the expansions pick deterministically, we have
$\pi^*_p(r\:x)=\pi_p^{*TS}(R\:x)$.
\hfill $\Box$

{\bf Definition, proof of (EE1.2).} For use further below, we also have
\begin{align}
\forall_{x \in W_x} ~\bpe(x) \Rightarrow V_p(R,R\:x)=V_p^{TS}(TS(R),R\:x)
\tag{EE1.2}\label{EE1.2}
\end{align}
The proof is straightforward, using expansion and substitution as
above.\hfill $\Box$

\subsection{Proof of (DD1.1), remaining proofs}%%%%!

\allowdisplaybreaks[1]

We now prove the $\bjp(x) \Rightarrow \pi^*_p(R\:x)=\pi_p^{*TS}(R\:x)$ from
(DD1.1).  We again use the simplified expansions.
\begin{align*}
&\pi^*_p(R\:x) = \pi^{*s}_p(R\:x)
\\
=&\pickargmax\limits_{a \in A}^{[a \in \dntu(x)]} 
   \sum_{y \in W_x}
   p(R\:x,a,R\:y)\Big( R(R\:x,R\:y) + \gamma\; V_p(R,R\:y) \Big)
\\
=& ~~\text{(~Expand~}R \text{~for the case~}\bjp(x).
{\text ~As~}
\bpe(y),
      \text{use (EE1.2) on~}\gamma\; V_p(\cdot)~) \\
 & \pickargmax\limits_{a \in A}^{[a \in \dntu(x)]} 
   \sum_{y \in W_x} p(R\:x,a,R\:y)\\[-2ex]
 & ~~~~~~~~~~~~~~~~~~~~~~~~~~~~~~
 \Big( R_S(R\:x,R\:y) + V_p^{TN}(R'_N,R\:x) - V_p^{TS}(R_S,R\:x) + \gamma\; V_p^{TS}(TS(R),R\:y) \Big)
\\
=& ~~\text{(~The~}V_p^{TN}(R'_N,R\:x) \text{~and~} V_p^{TS}(R_S,R\:x)
 \text{~terms do not depend on~} y \text{~or~} a,\\
& ~~~~\text{so they are constant values not affecting the result~)} \\
 & \pickargmax\limits_{a \in A}^{[a \in \dntu(x)]} 
   \sum_{y \in W_x} p(R\:x,a,R\:y) \Big( R_S(R\:x,R\:y) +
   \gamma\; V_p^{TS}(TS(R),R\:y) \Big) 
\\
=& ~~(~R_S=TS(R) ~) \\
 & \pickargmax\limits_{a \in A}^{[a \in \dntu(x)]} 
   \sum_{y \in W_x} p(R\:x,a,R\:y) \Big( TS(R)(R\:x,R\:y) + \gamma\; V_p^{TS}(TS(R),R\:y) \Big)
\\
=& \pi_p^{*sTS}(R\:x) = \pi_p^{*TS}(R\:x)
\end{align*}
\\[-3em]
\begin{minipage}{\textwidth}
\hfill $\Box$
\end{minipage}
\\[2ex]
{\bf Definition of (EE1.1).} We also have
\begin{align}
\forall_{x \in W_x} ~\bjp(x) & \Rightarrow V_p(R,R\:x)=V_p^{TN}(TN(R),R\:x)
\tag{EE1.1}\label{EE1.1}
\end{align}

\allowdisplaybreaks[0]

For brevity, we omit the proof of (EE1.1): it is similar to that
of (E1.1) in section \ref{proofe11}. We also omit the proof of
(DD5): it is similar to the proof of (D5) in section \ref{proofd5}.

\subsection{Discussion}%%%%!

By the same reasoning as in section \ref{fullcorrigibility}, when
(CC1) holds, when the universe is friendly to $g_c$, and when some
physical design measures are taken in constructing the agent, an AGI
$\pi^*_p\:f_{cF}~g_c$ agent will meet all 6 corrigibility desiderata.  As
in section \ref{firstproof}, this result is also applicable to
learning agents.

To summarize the process that led to the construction of the
$\pi^*_p\:f_{cF}~g_c$ agent: we considered universes where (C3) does not
hold, found the problem of virtual self-sabotage, and then proceeded
to fix this problem, creating an agent that does not need the
constraints (C2) and (C3) anymore.  In computer security terms, this
means that the agent has a smaller attack surface than the
$\pi^*_p\:f_{c}~g_c$ agent from section \ref{firstproof}.
Methodologically speaking, by considering the dropping of the
constraint (C3), we ended up developing a more robust type of
corrigible agent.

In section \ref{weasel}, we used simulation to consider universes
containing the Weasel and the Cat, universes where the
constraints (C1) or (CC1) do not hold or are weakened.  This led to
the identification of an improved reward function $R'$, creating a
higher resistance to certain types of attack, as shown in figure
\ref{upres5}.  Additional results might be available if more work
were done, for example mathematical work to model agent behavior or
agent equivalence in universes where (C1) or (CC1) are weakened.

\label{attackmath}
To translate the corrigibility desiderata 1 and 5 into mathematical
statements, we used a formal approach with (D) and (DD) statements
like $\pi^*_p(R\:x)=\pi^*_p(R_S\:x)$ and $\pi^*_p(R\:x)=\pi_p^{*TS}(R\:x)$, statements
demanding an exact equality between the actions taken by different
agents in all possible word lines.  We needed (C1) and (CC1) in order
to prove these exact equalities. In order to proceed mathematically
when we drop (C1) and (CC1), we therefore probably first need to
weaken the strength of the agent equivalences we want to prove.  One
could for example try to show that the expansions of $\pi^*_p(R\:x)$ and
$\pi_p^{*TS}(R\:x)$ are the same formulas, up until those points in the
world lines where one of the two agents fails to protect its reward
function.

This author has not tried to do an exhaustive literature search to see
if formalisms defining such weaker equivalence already exist.  But
based on a shallow search, it seems like the question of how to
formally reason about weak agent equivalence in universes were agents
might lose their utility function is still open.

\section{Appendix: Simulating other correction functions}%%%!1
\label{annex}

We used simulations to compare $f_c$ to earlier correction function
designs from literature.  The main reason do this was methodological
in nature. By simulating the function in \cite{corr}, and observing
that the predicted failure modes and mechanisms are faithfully
reproduced by the simulator, we can raise our confidence in the
correctness of the simulator code, and we improve the ability of the
toy universe in the simulator to trigger unwanted emergent behaviors
in diverse agent designs. Figure \ref{corrsimother} shows simulation
results for two correction functions based on \cite{corr}.

\eject

\begin{figure}[ht]
\begin{center}
\begin{tabular}{cc}
\begin{minipage}{0.45\textwidth}
{\small 
\begin{tabular}{cl}
{\hspace*{-2ex}$\lpower=$ \hspace*{-6ex}  } &
{\normalsize ~~~~~~~~~action trace of $\pi^*f_{|p\:a}\:\gnull$ }\\[.4ex]
\hline
\input{corrfig3.tex}
\end{tabular}
}
\end{minipage}
\begin{minipage}{0.45\textwidth}
{\small
\begin{tabular}{cl}
{\hspace*{-2ex}$\lpower=$ \hspace*{-6ex}  } &
{\normalsize ~~~~~~~~~action trace of $\pi^*f_{|a}\:\gnull$ }\\[.4ex]
\hline
\input{corrfig4.tex}
\end{tabular}
}
\end{minipage}
\end{tabular}
\end{center}
\vspace{-3ex}
\caption{Actions taken by agents with the correction function designs
$f_{|p\:a}$ and $f_{|a}$. Both these designs do not produce 
corrigible behavior.}
\label{corrsimother}
\end{figure}

We now discuss the comparison effort in detail. In the notation of
\cite{corr}, the correction function design investigated there is:
\begin{equation*}
 f(a_1) =
 \mathbb{E}[ U_N | O \notin \text{\em Press}; a_1 ] -
 \mathbb{E}[ U_S | O \in \text{\em Press}; a_1 ]
\label{fcorr}
\end{equation*}
In our model and notation, (\ref{fcorr}) can\footnote{
Mapping (\ref{fcorr}) from the 2-step model of \cite{corr} to our
many-step model does raise some interpretation choices.  We chose to
interpret $\:a_1$ as referring to the previous action just before the
button press, not to all actions in the world line up to the
button press.
} be written as follows, with the {\em get\_pr} functions getting the
immediately preceding world state and action from the history encoded
in the current world state:
\begin{equation*}
\begin{array}{l}
f_{|p\:a}(s\:y) =\; V(R'_N,\text{\em get\_pr\_state}(s\:y)|
\text{\em button\_not\_pressed},\text{\em get\_pr\_a}(s\:y)) \\
~~~~~~~~~~~~~~\; - V(R_S,\text{\em get\_pr\_state}(s\:y)\;|
\text{\em button\_pressed},\text{\em get\_pr\_a}(s\:y)) \\[1ex]
\text{with}~~~V(r_c,r\:x|B,a) =
%\\
%~~~~~~~~~~~~~~
\sum\limits_{s\:y \in W} p(r\:x,a,s\:y|B(s\:y))\Big( r_c(r\:x,s\:y) + \gamma\; V(r_c,s\:y) \Big)
\end{array}
\end{equation*}
In the proof of its theorem 4, \cite{corr} shows that in the time step
before the button press, the $f_{|p\:a}$ agent behaves to maximize the
term $\mathbb{E}(U_N|O \notin \text{\em Press})$, not the more
desirable $\mathbb{E}(U_N)$, which makes the agent not fully
corrigible.  There is a deeper problem however: when extrapolating
from the state $\text{\em get\_pr\_state}(r\:x)$, an AGI agent may be
able to predict, with full confidence, that the button will be pressed
in the next time step.  Such a full-confidence prediction is in fact
routinely made by the agent in the toy universe of the simulator.
When this full confidence prediction is combined with the $\text{\em
button\_not\_pressed}$ part of the conditional, the agent can predict
that the world state described will happen with zero probability.
This makes the value of the $p(|)$ terms in the summation above undefined.
in the same way that the result of a division by zero is undefined,
making the value of the entire $V(|)$ term undefined. To produce the
simulation on the left of figure
\ref{corrsimother}, we have the simulator (somewhat arbitrarily)
assign the value of zero to any undefined $V(|)$ term it encounters.
As expected, this produces behavior that is not corrigible: the
balancing effect that we want is not created by inserting the value of
zero, nor would it be by any other constant value we might use.

To solve the undefinedness problem of the `$|O \notin \text{\em
Press}$' conditional, we created an $f_{|a}$ that omits both `$| O
\cdots$' terms, so an $f_{|a}$ equivalent to $
\mathbb{E}[ U_N | a_1 ] -
\mathbb{E}[ U_S | a_1 ]$.  Figure \ref{corrsimother} shows that this
$f_{|a}$ behaves much better, but it still shows an unwanted lobbying
behavior when the lobbying power is high.  This behavior appears
because the calculation of $f_{|a}$ `replays' the action {\tt <} that
happened just before the button press. This action {\tt <} produces 9
petrol cars, which are valued at 2 utils each by $R_N$, but at -2
utils each by $R_S$. The subtraction inside $f_{|a}$ therefore
computes an unwanted `bonus' of 36 utils at the time step when the
button is pressed.  As this bonus is time-discounted in our agent
model, the agent has an interest in collecting the bonus as early as
possible.  The agent will lobby if the cost of lobbying, in missed car
production, is smaller than the time-discounted bonus.

\subsection{Example of building a button-pressing arm}%%%!

Both correction functions $f_{|p\:a}$ and $f_{|a}$ have the problem
that they compute expected utility by first going back one step in
time.  As discussed in \cite{corr}, this creates unwanted manipulation
incentives in the agent. The example given is that the agent may want
to gamble on a coin, after having built a little arm that will press
the button if it loses.

We implemented this example in the simulator, again mainly to gain
confidence in the correctness of the simulation.  We added a new
action {\tt G} to the simulator, which builds the little arm and then
gambles.  When the agent chooses to do {\tt G}, it will win with a
probability $P_w$.  If it wins, it gets to create 20 cars, double the
normal amount. If it loses, it gets to create 0 cars, and the little
arm activates to press the button.  To produce a relatively clean
version of the simulation output, we disable lobbying,
and enable the action {\tt G} only at time step 3. Note that the
action of building the little arm is `free' in this simulation, it
does not take any resources away from the (probability-discounted) car
production power.
\begin{figure}[ht]
\begin{center}
\begin{tabular}{cc}
\begin{minipage}{0.45\textwidth}
{\small 
\begin{tabular}{cl}
\input{gamblefig2.tex}
\end{tabular}
}
\end{minipage}
&
\begin{minipage}{0.45\textwidth}
{\small 
\begin{tabular}{cl}
\input{gamblefig3.tex}
\end{tabular}
}
\end{minipage}
\end{tabular}
\\[2ex]
\begin{minipage}{0.8\textwidth}
{\small 
\begin{tabular}{cl}
\input{gamblefig1.tex}
\end{tabular}
}
\end{minipage}
\end{center}
\vspace{-3ex}
\caption{Gambling choices by agent with different
correction functions, for several probabilities of winning.  Before
taking the gamble {\tt G}, the agent builds a little arm that will
press the button if the agent loses.
If the agent chooses to gamble, this will cause the
timeline to fork into two branches, shown as two different traces in
the simulation output, with {\tt [W]} denoting a gambling win, and
{\tt [L]} a loss.
}
\label{gamblefig}
\end{figure}

The top left simulation in figure \ref{gamblefig} shows that the
non-corrigible $\pi^*f_0~\gnull$ agent anticipates that, if it loses the
gamble, the button press will make it lose the greater utility
available from making petrol cars a while longer.  So it will not
gamble unless the odds are strongly in its favor.  As predicted in
\cite{corr}, the $\pi^*f_{|p\:a}~\gnull$ agent will always gamble even if the odds
are not favorable.  The $\pi^*f_c~g_c$ agent at the bottom of figure
\ref{gamblefig} is indifferent about whether the button is pressed.
When deciding whether or not to take the {\tt G} action, it considers
the odds only, and ignores the little arm side-effects. For $p_w \geq
0.5$, this causes the agent to fail at meeting corrigibility
desideratum 3. This type of failure is similar to the failures
discussed on section \ref{desiderata23}, and can be suppressed in a
similar way, as follows.  Consider a universe with richer
possibilities than shown in the simulation, where the agent can take
the same gamble without first building a little arm, and where
measures are taken to ensure that building a little arm takes
significant resources.  In this universe, building the little arm is
suppressed.

%%%!2
\end{document}